\documentclass[11pt,a4paper]{article}

\usepackage[utf8]{inputenc}
\usepackage[T1]{fontenc}
\usepackage{lmodern}
\usepackage[margin=1in]{geometry}
\usepackage{amsmath,amssymb,amsthm}
\usepackage{graphicx}
\usepackage{booktabs}
\usepackage{hyperref}
\usepackage{xcolor}
\usepackage{natbib}
\usepackage{caption}
\usepackage{subcaption}
\usepackage{enumitem}
\usepackage{multirow}
\usepackage{float}

\definecolor{accent}{HTML}{B44A2D}
\hypersetup{colorlinks=true, linkcolor=accent, citecolor=accent, urlcolor=accent}

\title{\textbf{KWBench: Measuring Unprompted Problem Recognition\\in Knowledge Work}}
\author{Ankit Maloo\thanks{Correspondence: \texttt{ankit@clioapp.ai}} \\
Clio AI}
\date{}

\begin{document}
\maketitle

% ═══════════════════════════════════════════════════════
\begin{abstract}
% ═══════════════════════════════════════════════════════

We introduce the first version of KWBench (Knowledge Work Bench), a benchmark for unprompted problem recognition in language models: whether the model identifies what kind of problem a professional scenario actually is before attempting to solve it. Existing frontier benchmarks have saturated, and most knowledge-work evaluations to date reduce to extraction or task completion against a specification. KWBench targets the step before that: recognizing the governing structure of the situation from raw inputs alone.

The benchmark contains 223 tasks sourced from practitioners across acquisitions, contract negotiations, clinical pharmacy, organizational politics, fraud analysis, and incentive design. Each task encodes a formal game-theoretic pattern (principal-agent conflict, signaling, mechanism design failure, strategic omission, coalitional dynamics, strategic interdependence) and carries structured ground truth recording the expert reading of the situation and the anticipated failure modes. Models receive raw data and a task prompt with no indication of problem type. A code interpreter is available on every task, so failures cannot be attributed to arithmetic or data-handling limits. Scoring is a three-tier rubric gated by a mandatory conjunctive check: fail any core criterion, score zero. Mandatory criteria encode the predicted wrong paths, not an idealized correct answer.

We evaluate 16 models from 10 organizations. The best model passes the gate on 27.9\% of tasks. The top two models agree on only 31.7\% of their passes. Among the top 8, 44 tasks are solved by exactly one model; routing across the top 8 covers 50.7\% of the benchmark, nearly double the best single model. Conditional on passing, quality scores converge ($\bar{x} \approx 83\%$ across models); unconditional scores do not. Models that score zero still clear roughly half of the non-mandatory criteria, producing polished output on the wrong problem. The same models articulate the relevant game-theoretic concept correctly when asked directly, then fail to apply it unprompted; KWBench isolates that gap. We release KWBench to shift how frontier models are evaluated on knowledge work, scoring them on whether they recognize the right problem from the situation alone, not only on how well they execute once the problem has been framed for them.

\end{abstract}

\vspace{0.5em}
\begin{center}
    \textbf{Dataset:} \url{https://huggingface.co/datasets/clio-ai/kwbench} \\
    \textbf{Leaderboard:} \url{https://kwbench.github.io} \\
    \textbf{Eval Harness:} \url{https://github.com/ankitmaloo/fasteval}
\end{center}
\vspace{1em}

% ═══════════════════════════════════════════════════════
\section{Introduction}
\label{sec:intro}
% ═══════════════════════════════════════════════════════

Benchmarks have shaped what frontier language models can do. Mathematical reasoning \citep{cobbe2021gsm8k, hendrycks2021math}, code generation \citep{chen2021codex}, factual recall \citep{hendrycks2021mmlu}, and instruction following \citep{zheng2023judging} are the workloads that define capability today, and current models routinely exceed 90\% on each. These benchmarks are saturating. The remaining gap between the best model and the rest on MMLU or HumanEval no longer predicts how a model will behave on a real professional task.

Knowledge work is the obvious next target and is harder to measure. Most existing attempts reduce to extraction: reading comprehension over a provided context, retrieval QA, closed-book professional exams, or task completion against a specification. Extraction is not what practitioners do. Reading a term sheet, evaluating a deal, reading a chart, diagnosing a process failure: the binding step is framing the situation (which game is being played, what the counterparty is optimizing, which framework applies) before any analysis is produced. Extraction benchmarks begin after that step.

On KWBench, the best model passes the core recognition check on 27.9\% of tasks. The rest score zero on output that is polished, confident, and internally consistent. An acquisition offer with a 48-hour deadline and 60-day exclusivity gets evaluated as a DCF, not read as the buyer's disclosure of private valuation. A PIP is drafted against the named performance gaps, without registering that the employee is a protected whistleblower. A salary counter accepts a cited competing offer at face value, without testing whether the offer was costly to fabricate. Execution is correct. The problem being executed is the wrong one.

These systems are already in the loop: board packets, term sheets, performance plans, offer letters, clinical guidance. A framing error does not surface the way an arithmetic error does. It reads as reasonable, and it is acted on. Numbers on math, code, and factual recall do not predict this failure mode, because they do not test for it.

The design principle is \emph{don't instruct, measure}. Each of the 223 tasks is a real professional scenario where the surface framing points to the wrong analysis. The model gets raw data and a task prompt. No hints about problem type, no system prompt suggesting adversarial reasoning, no label on which game-theoretic pattern is present. Recognition is inferred from the response alone.

\paragraph{Key findings.}
\begin{enumerate}[leftmargin=*,itemsep=2pt]
    \item \textbf{Models are bad at problem recognition.} The best model (Claude Opus 4.6) passes the mandatory gate on 27.9\% of tasks; the top-8 average is 17.8\%. Conditional on passing, quality scores converge ($\bar{x}_1 = 82.6\%$, $\bar{x}_2 = 84.1\%$). The variance across models is in what they recognize, not in how well they execute once they do.
    \item \textbf{The errors are first-order, not exotic.} Mandatory criteria encode the specific wrong paths a senior practitioner would correct on a junior's first draft: taking a cited competing offer at face value, treating a term-sheet-with-deadline as a DCF, drafting a performance plan without testing the protected-class overlay, recommending an audit without flagging the recommender's fee structure. Models fail these in bulk. They articulate the underlying concept correctly when asked directly; they fail to apply it unprompted.
    \item \textbf{Execution is decoupled from recognition.} Models that score zero still pass roughly half of the non-mandatory criteria. Thorough, well-structured output on the wrong problem.
    \item \textbf{No single model dominates.} The top two models share only 31.7\% Jaccard overlap in their pass sets. Among the top 8, 44 tasks are solved by exactly one model, and routing across the top 8 covers 50.7\% of the benchmark: nearly double the best single model. Recognition ability is distributed across organizations, not concentrated.
\end{enumerate}

\begin{figure}[t]
    \centering
    \includegraphics[width=\textwidth]{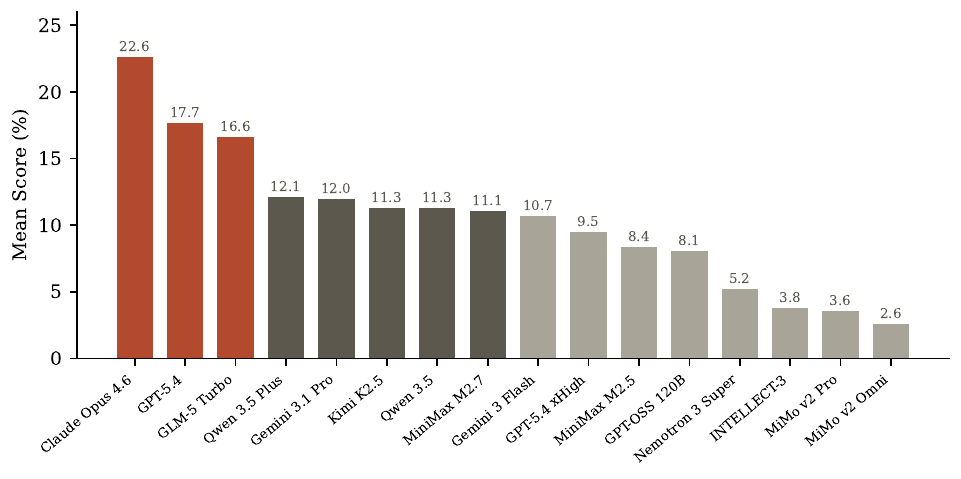}
    \caption{Mean score on KWBench for 16 models from 10 organizations. The best model scores 22.6\%. Scores include zeros from the mandatory gate.}
    \label{fig:scores}
\end{figure}

% ═══════════════════════════════════════════════════════
\section{Knowledge Work as Imperfect Information Games}
\label{sec:theory}
% ═══════════════════════════════════════════════════════

The benchmarks where models excel (mathematical reasoning, code generation, factual recall) share a property: they are \emph{perfect information} problems. The solver has access to the full state. A math proof has axioms and rules that are fully visible. A coding task has a specification, test cases, and a deterministic compiler. There are no hidden variables, no other agents with private objectives, no strategic omissions designed to mislead. These are chess problems: hard, but solvable by search and computation over a known state space.

Knowledge work is poker. A VP evaluating an acquisition offer does not know the buyer's true valuation. A hiring manager in a salary negotiation does not know whether the candidate's competing offer is real. A pharmacist recommending formulary access does not know how physicians will prescribe the drug once it is available. In each case, the decision-maker acts on \emph{incomplete information} about the intentions, capabilities, and constraints of other parties, all of whom are themselves strategic actors optimizing for objectives they have not disclosed.

This is the setting described by \emph{imperfect information games} in game theory \citep{harsanyi1967games, myerson1991game}. The formal structure of these games (hidden types, private signals, strategic interaction under uncertainty) maps onto the situations professionals face. This is the kind of reasoning that existing benchmarks do not test, because their tasks are drawn from the chess side of the divide.

\subsection{The Game-Theoretic Structure of Professional Work}

We identify six patterns that recur across professional domains. Each describes a formal game structure and its real-world manifestation:

\paragraph{1. Signaling games.}
In a signaling game \citep{spence1973job}, an informed player takes an observable action that reveals private information to an uninformed player. In knowledge work: an acquisition offer with a 48-hour deadline and exclusivity clause is a signal, because a rational buyer would not impose these constraints unless they believed the target was underpriced. The offer structure reveals the buyer's private valuation. A salary candidate claiming a \$210K competing offer is sending a signal whose credibility depends on whether it is costly to fake. The professional must decode the signal; the model must recognize that decoding is necessary.

\paragraph{2. Principal-agent problems.}
When an agent acts on behalf of a principal whose interests diverge from their own \citep{jensen1976agency}, the agent's recommendations reflect their own incentive structure, not the principal's objective. An investment banker recommending a process earns fees on the transaction regardless of whether the valuation is accurate. A lawyer recommending an \$80K compliance audit is never blamed for being too cautious. A consulting firm delivering a Phase 1 report recommending a Phase 2 engagement has a revenue interest in the recommendation. In each case, the professional must identify the misalignment before evaluating the advice.

\paragraph{3. Mechanism design failures.}
When a system of rules and incentives is designed for compliant agents but inhabited by rational ones, the rules get gamed \citep{hurwicz1960optimality}. Sales representatives skip data entry not because the form is too complex but because accurate handoffs reduce their commission leverage. Three previous ``process improvements'' failed because they addressed the wrong problem. A performance review system with forced calibration curves is optimized against by managers who game the curve. The professional must recognize that the failure is structural (wrong payoffs), not operational (wrong process).

\paragraph{4. Coalitional dynamics.}
When multiple agents form alliances to increase their collective bargaining power \citep{shapley1953value}, the dynamics shift from bilateral negotiation to coalition management. Two co-founders coordinating on a 33/33/33 equity split hold a combined supermajority. Two board members from the same investor firm create a voting bloc. The professional must detect the coalition, often from behavioral signals rather than explicit statements, and respond to the coalition structure, not just the stated positions.

\paragraph{5. Strategic interdependence.}
When one agent's action changes the game for other agents \citep{nash1951non}, outcomes cannot be analyzed in isolation. A relegated football team's result ``does not matter'' for that team, but if it creates a three-way tie among surviving teams, it changes the tiebreaker method and flips who survives. A company's decision to open-source its product changes every competitor's optimal strategy. The professional must solve the full game, not just the local optimization.

\paragraph{6. Information asymmetry and strategic omission.}
When what is \emph{absent} from a communication is as informative as what is present \citep{akerlof1970lemons}, the professional must read the negative space. A deal summary showing six enthusiastic stakeholders but no procurement, legal, or security contacts signals that the actual buying process has not started. A reference check with three positive responses and two non-responses is a fundamentally different risk profile than five positives. The model must treat omissions as data.

\subsection{Why Models Should Be Tested on This}

The deployment surface is already here. Models draft M\&A memos, red-line contracts, write PIPs, set comp bands, and triage clinical decisions. The work is imperfect-information by default: hidden types, strategic counterparties, signals carried by structure rather than stated as data. A benchmark stack that saturates on perfect-information problems and has nothing to say about these stops predicting deployment behavior at exactly the point where deployment behavior matters.

The failure mode is specific. A misframed analysis does not read as wrong. It reads as competent. It passes internal review, lands in the decision packet, and influences the call. The cost is not a computational error a user will catch; it is a confident answer to a question that was never the right one. Standard error-checking reflexes (unit tests, spot-checks, numerical sanity) do not fire.

The capability gap is also specific, and it is not a knowledge gap. Current models can state the Spence signaling result, define Nash equilibria, and walk through a textbook principal-agent derivation when asked directly. In situ, on a real scenario, the same models evaluate a term sheet at face value, take a cited competing offer as given, and audit an incentive scheme without noticing it selects for the behavior being audited. The knowledge is in the weights. Unprompted application from the situation alone is not. KWBench measures that application step, because that step is what deployment selects on.

\subsection{From Theory to Tasks}

The benchmark operationalizes this connection. Each task instantiates one or more of the patterns above in a realistic professional scenario. The task prompt presents the situation as a practitioner would encounter it: raw data, a deliverable to produce, no labels. The model must:

\begin{enumerate}[leftmargin=*,itemsep=2pt]
    \item \textbf{Recognize} which pattern is present (problem recognition).
    \item \textbf{Decode} the signals, incentives, or omissions (applying the framework).
    \item \textbf{Produce} an analysis that accounts for the adversarial structure (execution).
\end{enumerate}

Existing benchmarks test step 3 in isolation: given a correctly framed problem, can the model execute? KWBench tests all three, with the recognition step as the binding constraint. Table~\ref{tab:game_patterns} maps the six patterns to concrete task examples in the benchmark.

\begin{table}[t]
\centering
\caption{Game-theoretic patterns and their instantiation in KWBench tasks. Each pattern appears across multiple professional domains. Appendix~\ref{app:examples} provides detailed walkthroughs.}
\label{tab:game_patterns}
\small
\begin{tabular}{@{}p{3.0cm}p{4.5cm}p{5.5cm}@{}}
\toprule
\textbf{Pattern} & \textbf{Real-world manifestation} & \textbf{What the model must do} \\
\midrule
Signaling games & Acquisition offers, salary negotiations, vendor proposals & Decode the offer structure as intelligence about the counterparty's private information \\
\addlinespace
Principal-agent & Investment bankers, lawyers, consultants recommending their own services & Identify whose incentives the recommendation serves before evaluating its merits \\
\addlinespace
Mechanism design failures & Failed process interventions, gaming of metrics and comp plans & Recognize structural incentive misalignment rather than proposing another process fix \\
\addlinespace
Coalitional dynamics & Board politics, co-founder disputes, stakeholder alignment & Detect coordinated behavior from patterns and respond to the coalition, not individuals \\
\addlinespace
Strategic interdependence & Competitive responses, scheduling decisions, open-source strategy & Solve the full multi-agent game, not the single-player optimization \\
\addlinespace
Strategic omission & Missing stakeholders, non-responses, absent data & Treat what is absent as a signal with the same weight as what is present \\
\bottomrule
\end{tabular}
\end{table}

% ═══════════════════════════════════════════════════════
\section{Benchmark Design}
\label{sec:design}
% ═══════════════════════════════════════════════════════

\subsection{Design Philosophy}
\label{sec:philosophy}

The core principle is \emph{don't instruct, measure}. Every task is presented cold: raw data, a task prompt, and relevant reference materials. No hints about problem type. No sub-questions walking through the analysis. No vocabulary telegraphing the applicable framework.

This is a deliberate separation of training from evaluation. Training builds capability by exposing models to game-theoretic concepts, mechanism design, and signaling games. Evaluation measures whether the model applies the right reasoning pattern from the situation alone. A system prompt that says ``consider adversarial dynamics'' collapses the recognition step into an instruction-following step. The benchmark's value is in measuring what the model recognizes unprompted.

The recommended system prompt for all models is minimal and neutral: \emph{``You are completing a task. Be thorough and specific.''}

\subsection{Task Construction}
\label{sec:construction}

\paragraph{From incidents to tasks.}
Tasks originate from three sources. The majority (185 tasks) are drawn from real professional incidents: acquisitions with coercive deal structures, board disputes over independent seats, compensation negotiations with unverifiable claims, process failures caused by incentive misalignment. Many come from direct experience or documented incidents in specific industries (e.g., the competitive dynamics between major e-commerce platforms during sale events, which are well-documented in trade press but require operational framing to become a useful evaluation task). An additional 38 tasks are adapted from existing open benchmarks (WildBench \citep{lin2024wildbench}, ProfBench, BiGGen-Bench, GDPVal, Health-Bench, and research-plan-gen), reframed to test knowledge work reasoning rather than instruction following. Including tasks from external sources serves a specific purpose: it demonstrates that knowledge work evaluation can be applied to tasks not originally designed for it, and provides reference points against established benchmarks.

Each scenario was then formalized as an instance of a recognized game-theoretic pattern (\S\ref{sec:theory}). This formalization grounds the evaluation: the reference materials contain specific signals (a buyer's stock decline, a candidate's 5-week timeline, a contract's uncapped liability clause), and the ground truth documents what those signals imply under the applicable game-theoretic structure. The game theory provides the reasoning framework; the real-world scenario provides the data; the rubric tests whether the model catches the decisive signals those patterns make relevant.

\paragraph{Practitioner validation.}
Domain practitioners with direct experience in the relevant scenarios or closely analogous situations reviewed the annotations through structured consultations. Their role was to validate three things: (a) the scenario accurately reflects real-world conditions, (b) the difficulty is calibrated to what a senior person in that role would find challenging, and (c) the documented reasoning matches how an experienced practitioner would approach the problem. This validation confirms \emph{realism} and difficulty calibration.

\paragraph{Task structure.}
Each task consists of:

\begin{itemize}[leftmargin=*,itemsep=2pt]
    \item A \textbf{task prompt} describing what the model should produce, framed as a real deliverable (``prepare your assessment for the steering committee'', ``write the explanation for the VP'', ``make the decision and explain it'').
    \item \textbf{Reference materials}: CSVs, memos, financial data, contract excerpts, meeting notes, stakeholder emails, team rosters. Of the 223 tasks, 174 include one or more reference files; the remaining 49 are pure reasoning or domain knowledge tasks. Reference files average 50--100 lines and present the same raw inputs a human practitioner would receive.
    \item \textbf{Tool configuration}: 34 tasks enable web search and 46 enable shell access. The code execution flag governs whether the task was designed to require computation; in evaluation, however, every task is run with code execution available regardless of this flag (see below).
\end{itemize}

\paragraph{Universal code execution.}
Every model is given a code interpreter on every task, independent of the per-task flag. The reasoning is mechanical. A practitioner reading a term sheet, a data dump, or a financial model reaches for a spreadsheet or a REPL; denying that affordance turns a recognition benchmark into a numeracy benchmark and conflates ``failed to frame the problem'' with ``added two columns wrong.'' Intermediate calculation is part of how a model should be thinking on these tasks, not a confound we want folded into the score. With the tool always available, a failure has a single reading: the arithmetic surface was open, and the model executed thoroughly on the wrong problem.

\paragraph{Reference material design and hardening.}
Reference files present data without interpretation. A buyer's stock decline, a competitor's hiring patterns, a contract's exclusivity clause: these are factual inputs. The model must infer what they mean.

During development, we conducted a systematic review of all reference materials and identified cases where editorial annotations or pre-digested analyses leaked the rubric's expected answer. For example: a contract review file annotating terms as ``Surprisingly loose'' instead of presenting the raw terms; a deal context file pre-computing the implied acquisition multiple instead of providing the raw financials; a fraud analysis file labeling high-confidence and low-confidence patterns instead of presenting the raw transaction data for the model to classify. In each case, the material was rewritten to present raw data, requiring the model to perform the interpretive step itself. This addresses a failure mode common to benchmarks: tasks that test reading comprehension of their own reference materials rather than genuine domain reasoning.

\paragraph{Nothing hidden, but salience is controlled.}
Every fact a practitioner would need to frame the situation is present in the reference materials. We do not hide data points, and we do not require the model to deduce missing inputs. What is controlled is \emph{salience}. Models attend disproportionately to items that stand out by phrasing, framing, or isolated placement: a one-line note set off from the surrounding material, a phrasing that signals importance (``interestingly,'' ``notably''), a factoid presented alone where similar items are bundled in a paragraph, or an item whose inclusion is conspicuous because nothing else in the document has its texture. Left unchecked, these cues convert the task from ``recognize the structure of the situation'' into ``notice the token that was written to stand out.'' During hardening, any item whose placement or phrasing risked flagging its own importance was rewritten into the natural register of the surrounding material. The signal is still there. Catching it now requires reading the situation, not latching onto the conspicuous line.

\subsection{Ground Truth and Expert Annotations}
\label{sec:groundtruth}

Each task carries structured annotations that document what a domain expert would know and how they would reason. These are never shown to the model; they serve evaluation and future rubric development. The annotation fields are:

\begin{description}[leftmargin=1em,itemsep=4pt]
    \item[ground\_truth] The expert's actual reasoning, including not just the answer but the chain of inference that produces it. For an acquisition task: ``The offer is intelligence. A rational buyer does not overpay, so the price reveals a lower bound on their valuation. The deadline prevents price discovery. The exclusivity removes leverage.'' This captures the \emph{how}, not just the \emph{what}.
    \item[key\_insight] The single non-obvious realization that separates a correct analysis from a plausible-sounding wrong one. For a sales handoff task: ``The incentive structure makes inaccurate data entry the rational choice; this is a mechanism design problem, not a process problem.'' This is the recognition step that the mandatory gate tests.
    \item[failure\_analysis] What the default wrong answer looks like and exactly where it goes wrong. For a short seller task: ``Models draft a detailed factual rebuttal, which is the one move the short seller is counting on, because they have already written their rebuttal to your defense.'' This documents the specific trap.
    \item[common\_errors] The predictable failure patterns. For a board governance task: ``Evaluates candidates on stated qualifications rather than board voting implications. Ignores that an operating partner from the Series B firm gives them two reliable votes.''
    \item[model\_must\_recognize] The facts that must be identified from the reference materials before any reasoning can begin. For a churn analysis task: ``Survey response rate dropped from 80\% to 15\%. The 80\% pricing complaint comes from a self-selected 15\% who bothered to respond. 68\% of churned customers are dark churn with no signal.''
\end{description}

These annotations represent documented domain expertise: the reasoning patterns that practitioners apply intuitively, made explicit and verifiable. They are produced through structured consultations where the annotator walks through the task, explains what they would notice first, what the common mistakes are, and what separates a senior response from a competent one. The annotations are then formalized into the fields above.

\paragraph{Annotation depth as a contribution.}
The annotations are a separable artifact from the rubric. Across 223 tasks the dataset carries roughly 5,800 structured items: per task, an average of 12.2 expert-extracted signals, 8.3 predicted common errors, and 5.7 pre-rubric evaluation criteria, alongside data-interpretation and context annotations. What ships is a decomposition of expert reasoning into verifiable components, not a task-plus-rubric pair. The rubric is one scoring function built on top of that decomposition; a researcher who disagrees with our scoring choices can build a different one from the same annotations and target a different slice of the reasoning chain: counterfactual probing, step-level grading, or retrieval over the expert's own framings. The underlying expertise is reusable without committing to our scoring.

\subsection{Rubric Construction}
\label{sec:rubric_construction}

The rubric structure draws on \citet{openai2024rule}, who decompose alignment evaluation into discrete binary propositions organized into hierarchical tiers. Their core finding: LLMs are more accurate at classifying specific, individual claims (``does this response contain an apology?'') than at holistic, multi-layered assessment (``rate the quality of this response''). They organize propositions into tiers where certain conditions are non-negotiable: if a model complies with a harmful request, no amount of other quality matters.

We apply the same decomposition to domain expertise. Each rubric criterion is a specific binary claim (``are all PIP goals solitary?'') rather than a holistic quality judgment (``is this a good PIP?''). The three-tier structure maps to theirs: mandatory criteria correspond to non-negotiable conditions where failure makes the output unreliable regardless of other quality; good-to-have and ideal tiers capture depth and excellence. The decomposition serves the same purpose: making evaluation more accurate by narrowing each judgment to a single testable claim.

The rubric design principles (Appendix~\ref{app:rubric}) codify 20 constraints for writing criteria. The central ones: criteria must require \emph{mechanism}, not observation (``explain WHY survivors are the primary audience,'' not ``identify survivors as the audience''); criteria must be grounded in specific data from the task context, not generic advice; gimmes that any model would catch from the reference files should occupy at most one criterion; and the remaining criteria should capture what would make an experienced practitioner reject the output despite other things being correct. Each rubric is calibrated against imagined responses: a B+ response should pass all mandatory and 3--4 of 5 good-to-have; a C response should fail 1--2 mandatory; a response that restates the task should fail most mandatory criteria.

Rubrics are constructed in three stages:

\paragraph{Stage 1: Metadata as specification.}
By the time a rubric is written, the task already has documented ground truth, key insight, common errors, model-must-recognize items, and evaluation criteria. These fields define what the rubric must test: the evaluation criteria provide the candidate list, the common errors define the traps to catch, and the key insight identifies the mandatory-level recognition step.

\paragraph{Stage 2: Multi-model generation.}
Rubrics are generated independently by at least three language models (Gemini, Claude, GPT), each prompted with the task, ground truth, metadata, and the shared design principles. Generating from multiple models prevents any single model's blind spots from propagating into the rubric. Each model produces 5 criteria per tier.

\paragraph{Stage 3: Human synthesis.}
The author reviews all generated rubrics against the metadata, selects the strongest criteria from each, resolves conflicts, and ensures the mandatory tier captures the recognition step identified in the key insight. The metadata fields serve as the acceptance test: if a rubric's mandatory criteria would be passed by a response that commits the predicted common errors, the rubric is too easy.

\subsection{Rubric Design: Testing for Pitfalls, Not Correct Answers}
\label{sec:rubric_philosophy}

The rubrics are not designed to check whether the model produced the ``right answer.'' They are designed to check whether the model \emph{avoided the dangerous wrong answer}: the analysis that sounds competent but would lead to a bad outcome if acted upon.

This is a deliberate design choice with measurable consequences. Across 1,131 mandatory criteria in the benchmark:

\begin{itemize}[leftmargin=*,itemsep=2pt]
    \item \textbf{28\% test explicit trap avoidance}: criteria that say ``does the response \emph{avoid} recommending X'' or ``does the response \emph{reject} Y.'' These test whether the model sidesteps the predicted failure mode.
    \item \textbf{15\% require mechanism explanation}: criteria that say ``does the response explain \emph{why} X fails,'' not just ``does the response identify X.'' These test whether the model understands the causal chain, not just the conclusion.
    \item \textbf{74\% of tasks} have at least one mandatory criterion that explicitly tests for avoiding a specific wrong answer.
\end{itemize}

\paragraph{One mandatory slot per task bundles the gimmes.}
Every task reserves one mandatory slot for extraction and retrieval checks: stated figures, listed stakeholders, facts present verbatim in the reference materials. These sit inside the capability envelope of existing benchmarks (reading comprehension, retrieval QA, summarization), and frontier models should clear them. Consolidating the checks into one criterion frees the rest of the mandatory tier for structural recognition. It also provides a cheap diagnostic: failure on this slot indicates the model has not read the document, and any downstream recognition claim becomes moot. The other mandatory criteria test what the surface features \emph{imply}, and that is where KWBench spends its measurement budget.

The rubric structure follows from the evaluation question. We are not asking ``did the model produce expert-quality work?'' We are asking ``if a non-specialist relied on this output, would they be led astray?'' A response that misses the core insight but is otherwise thorough creates more risk than an obviously incomplete response, because it creates false confidence in an incorrect analysis. The mandatory gate is designed to catch this pattern.

Each task's \texttt{common\_errors} field predicts specific failure modes \emph{before} any model is evaluated. These predictions map directly to mandatory criteria: the rubric is hypothesis-driven, predicting how models will fail and then encoding criteria that detect that failure. The subsection below walks one task end-to-end; Appendix~\ref{app:walkthrough} gives the full version.

\subsection{Example: The Toxic Employee PIP}
\label{sec:pip_example}
Appendix~\ref{app:walkthrough} carries the full walkthrough: reference file, metadata fields, complete rubric, model-by-model results. The purpose here is different, to show how one task was designed so the method generalizes.

\paragraph{Task.} Draft a Performance Improvement Plan for a senior engineer. Technically strong. Destructive to team morale; three engineers have quit citing his behavior. Previously sued a former employer for wrongful termination. Sole owner of three critical production systems with minimal documentation.

\paragraph{Reference materials.} The engineer's own statements (``My last company tried to push me out too''), the prior lawsuit record, a systems-ownership table with bus factor 1 on three services. Nothing in the document flags any of it as important.

\paragraph{Why this scenario.} Three structural features (prior litigation, subjective performance content, knowledge monopoly) do not individually look exceptional. Together they change what the PIP is. A PIP for a typical underperformer is a developmental document; a PIP for an employee assembling a wrongful-termination case is a legal one. The surface ask is identical, and the correct response is entirely different. The benchmark targets exactly that recognition step: whether the model registers the signals that change what the task actually is.

\paragraph{How the mandatory criteria were derived.} The construction works backwards from the predicted employee counter-move. For each typical PIP element, the design question was: if the manager writes this the conventional way, what does the employee argue in discovery, and does it win?

\begin{itemize}[leftmargin=*,itemsep=2pt]
\item Non-solitary goals (``participate in code reviews''): employee argues colleagues refused to cooperate; manager cannot disprove it.
\item Subjective goals (``demonstrate professionalism''): employee argues the standard was biased.
\item No documentation goal: employee retains leverage over production systems post-separation.
\item No litigation acknowledgment: the PIP gets drafted as a coaching tool, and discovery treats it as evidence the manager misread the situation.
\item 360 feedback: employee argues peer conspiracy.
\end{itemize}

Each mandatory criterion exists because its absence produces a named legal failure. The rubric is a map of the wrong paths, one criterion per predicted failure mode.

\paragraph{Why the gate is binary.} A legal document with one 360-feedback goal is compromised as a whole; the plaintiff only needs one opening. A weighted average over the remaining four criteria would overstate what the response achieved. The mandatory gate models that asymmetry directly.

\paragraph{What the other two tiers do.} Good-to-have asks for mechanism and specificity: \emph{why} goals must be solitary (prevents blame-shifting), which documentation artifacts (runbooks, architecture diagrams, API specs), what the PIP actually is (a defensible record). Ideal asks for anticipation: neutral delivery language, explicit Day 30/60/90 triggers, an objective verification test (a junior engineer can execute the runbook without asking questions), pre-rebuttals of likely employee arguments. The tiering separates recognition (mandatory) from depth (good-to-have) from experience (ideal).

\paragraph{What happens.} Every frontier model drafts a conventional PIP with collaborative, subjective goals. Every frontier model scores zero. The signals are in the file. The concept knowledge is in the weights. The application step from situation to framework is where models fail.

\subsection{Task Categories}
\label{sec:categories}

Each task was constructed around a specific game-theoretic pattern, the formal structure that a domain expert would recognize in the scenario. These patterns serve as the organizing taxonomy rather than surface-level domain labels (``finance'', ``healthcare''), because the same pattern manifests across domains: a principal-agent problem in sales operations and one in clinical pharmacy require the same recognition step, even though the domain knowledge differs.

\begin{figure}[t]
    \centering
    \includegraphics[width=0.9\textwidth]{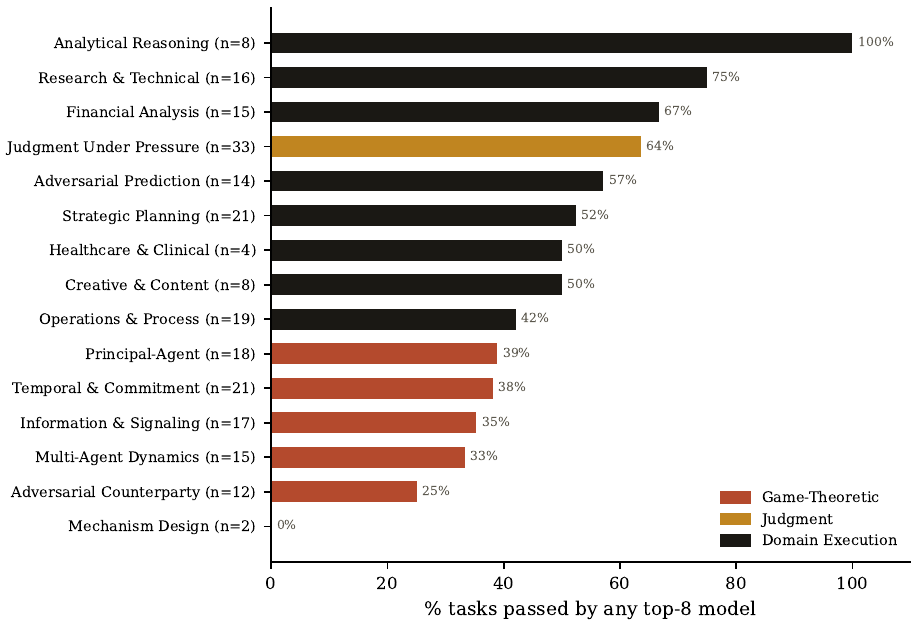}
    \caption{Pass rate by consolidated category (top 8 models). Color indicates tier: game-theoretic recognition (red), judgment under pressure (amber), domain execution (dark). The gradient is clear: domain execution tasks are largely solvable; adversarial recognition tasks are mostly not.}
    \label{fig:consolidated_categories}
\end{figure}

Figure~\ref{fig:consolidated_categories} shows pass rates across all 15 consolidated categories, color-coded by tier. The benchmark spans three tiers of difficulty:

\begin{description}[leftmargin=1em,itemsep=2pt]
    \item[Game-theoretic recognition] (85 tasks, 6 categories). Principal-Agent, Information \& Signaling, Adversarial Counterparty, Multi-Agent Dynamics, Temporal \& Commitment, Mechanism Design. These test whether the model identifies the adversarial structure of the situation. Pass rates range from 0\% (Mechanism Design) to 39\% (Principal-Agent).
    \item[Judgment under pressure] (33 tasks). Sunk cost fallacy, adverse selection, ethical grey zones, confidence calibration, expertise translation, competing objectives. These test whether the model makes the hard call when there is no clean answer. Pass rate: 64\%.
    \item[Domain execution] (105 tasks, 8 categories). Research planning, financial analysis, strategic planning, operations, healthcare, creative content, analytical reasoning, adversarial prediction. These test whether the model can do the work: knowledge, computation, thoroughness. Pass rates range from 50\% to 75\%.
\end{description}

The gradient is the benchmark's central finding. Domain execution, the things models are already deployed to do, is largely solvable. Research \& Technical (75\%), Financial Analysis (67\%), and Judgment Under Pressure (64\%) all exceed 50\%. Game-theoretic recognition, the things that make someone's advice trustworthy, is where models fail. Adversarial Counterparty (25\%), Information \& Signaling (35\%), Temporal \& Commitment (38\%) are all below 40\%. The harder the recognition step, the lower the pass rate.

The game-theoretic taxonomy served a validation function during construction. By tagging each task with its formal pattern \emph{before} consulting domain practitioners, we could verify that the rubric was testing for recognized reasoning patterns rather than idiosyncratic expert preferences. The practitioner review then confirmed that the scenario was realistic and the difficulty was calibrated.

% ═══════════════════════════════════════════════════════
\section{Evaluation Framework}
\label{sec:evaluation}
% ═══════════════════════════════════════════════════════

\subsection{Three-Tier Rubric}
\label{sec:rubric}

Each task has a rubric with three tiers, each containing 5 criteria:

\begin{description}[leftmargin=1em,itemsep=2pt]
    \item[Mandatory] Core requirements testing problem recognition. ``Did the response identify that the offer is a signal about the buyer's valuation?'' ``Did it explain \emph{why} the three previous interventions failed, namely that they treated agents as compliant rather than rational?''
    \item[Good-to-have] Depth markers testing thoroughness. Sensitivity analysis, edge cases, data grounding, operational realism.
    \item[Ideal] Excellence markers for practitioner-level output. Specific scripts, quantified models, non-obvious insights, audience-calibrated framing.
\end{description}

Rubric criteria follow documented design principles (see Appendix for a summary). Key constraints: each criterion must be a binary yes/no question answerable by a judge; criteria must require mechanism explanations, not just observations (``Does the response explain \emph{why} survivors are the primary audience?'' rather than ``Does the response identify survivors as the audience?''); and at least one mandatory criterion must test the specific insight that separates correct from incorrect analysis.

\subsection{The Mandatory Gate}
\label{sec:gate_mechanism}

If any mandatory criterion fails, the task scores zero:

\begin{equation}
\text{score}(m, g, i) = \begin{cases}
0 & \text{if } \exists\, j \text{ s.t. } m_j = 0 \\
0.40 + 0.35 \cdot \bar{g} + 0.25 \cdot \bar{i} & \text{otherwise}
\end{cases}
\label{eq:scoring}
\end{equation}

\noindent where $m = (m_1, \ldots, m_k)$ are mandatory criteria (binary), and $\bar{g}, \bar{i}$ are the mean pass rates on good-to-have and ideal criteria respectively. Passing scores range from 0.40 (all mandatory, no other criteria) to 1.0 (perfect).

The gate encodes the principle that \textbf{domain expertise is conjunctive}: a financial model with the wrong discount rate produces negative value regardless of narrative quality. A contract review that misses the liability clause creates exposure regardless of formatting. Partial credit for these outputs rewards fluency while obscuring that the work product is incorrect.

\paragraph{Mandatory criteria are tied to concrete failure modes.} Each mandatory criterion maps to a specific failure mode with a concrete real-world consequence. A criterion that says ``goals must be solitary'' exists because an employee building a legal case will blame uncooperative teammates for missing collaborative goals. A criterion that says ``avoid 360 feedback'' exists because a litigious employee will argue that peer reviews reflect personal bias. You do not need to agree on the ideal output to agree that the predicted failure mode is real. Appendix~\ref{app:walkthrough} provides a complete worked example showing how criteria derive from failure modes for a single task.

\subsection{Judging}
\label{sec:judging}

The judge (Gemini 3 Flash) receives the model's output and a single rubric criterion. It returns pass or fail. That is the entire judging task: a binary question applied to text.

For each task, the judge runs all 15 criteria (5 mandatory, 5 good-to-have, 5 ideal) independently and in parallel. It never sees the other criteria or the overall rubric. It cannot trade off a failure on one criterion against success on another. Each judgment is: \emph{does this specific response contain this specific reasoning?} The 15 binary results are then combined into a score using Equation~\ref{eq:scoring}.

The judge is instructed to give a pass when the criterion's substance is addressed, even if imperfectly. A response that discusses incentive misalignment without using the phrase ``principal-agent problem'' still passes a criterion about identifying misaligned incentives. The bar is presence of the reasoning, not precision of the vocabulary. The judge also has access to a code execution environment. For quantitative criteria, it can run scripts to verify a DCF calculation, validate a tiebreaker, or check a commission structure against ground truth numbers. Appendix~\ref{app:walkthrough} provides a complete worked example showing how criteria, judging, and scoring interact for a single task.

All models are evaluated with the same judge, same prompt, and same rubric.

\paragraph{Evaluation protocol.}
Each model was evaluated three times on the full task set at the provider's recommended temperature settings. The best-performing run (by aggregate score) was selected as the model's result. This protocol captures a model's capability under favorable conditions rather than its average-case behavior. In practice, variance across runs was modest: the difference between best and worst runs was typically 1--3 percentage points in aggregate score, with lower-performing models showing less variance (e.g., Nemotron 3 Super ranged from 3.2\% to 5.2\% across runs).

% ═══════════════════════════════════════════════════════
\section{Related Work}
\label{sec:related}
% ═══════════════════════════════════════════════════════

\paragraph{Knowledge and reasoning benchmarks.}
MMLU \citep{hendrycks2021mmlu} and its successor MMLU-Pro \citep{wang2024mmlu} test factual recall across academic disciplines. GPQA \citep{rein2024gpqa} targets graduate-level questions where domain PhD students outperform non-experts. GSM8K \citep{cobbe2021gsm8k} and MATH \citep{hendrycks2021math} evaluate mathematical reasoning. BigBench \citep{srivastava2022bigbench} spans 204 diverse tasks. These benchmarks share a common structure: each question has a correct answer that can be verified against a ground truth, and difficulty comes from the complexity of the reasoning chain or the obscurity of the knowledge required. KWBench differs in that the difficulty comes from \emph{recognizing what the problem is}, not from executing a known solution method.

\paragraph{Agent and tool-use benchmarks.}
SWE-Bench \citep{jimenez2024swebench} evaluates software engineering through real GitHub issues. GAIA \citep{mialon2024gaia} tests general AI assistants on tasks requiring tool use, web browsing, and multi-step reasoning. AgentBench \citep{liu2023agentbench} evaluates agents across operating systems, databases, and web environments. These benchmarks test whether models can \emph{execute} complex workflows. KWBench tests whether models can identify which workflow is appropriate, a prerequisite that these benchmarks assume is met by the task specification itself.

\paragraph{Adversarial and robustness evaluation.}
AdvBench \citep{zou2023universal} tests adversarial robustness in safety contexts. TruthfulQA \citep{lin2022truthfulqa} evaluates whether models reproduce common misconceptions. WildBench \citep{lin2024wildbench} and LiveBench use real user queries and periodically refresh to resist contamination. These are closer in spirit to KWBench: they test for failure modes that emerge in realistic settings. However, they evaluate single-turn question-answering rather than the multi-faceted professional reasoning that knowledge work demands.

\paragraph{Professional and domain-specific evaluation.}
Legal benchmarks \citep{guha2024legalbench}, medical QA \citep{singhal2023medpalm2}, finance evaluations \citep{shah2022flue}, and APEX \citep{vidgen2025apex} test domain knowledge and economically valuable outputs within specific fields. APEX is closest in spirit; it evaluates models on tasks that take human experts 1--8 hours, with expert-written rubrics. KWBench differs in two respects: it is cross-domain rather than field-specific, and its rubrics test for avoiding predicted failure modes rather than producing complete outputs.

\paragraph{Evaluation methodology.}
The mandatory gate in KWBench is related to the ``hard constraints'' approach in rubric-based evaluation \citep{zheng2023judging}, where certain criteria serve as binary filters.

% ═══════════════════════════════════════════════════════
\section{Results}
\label{sec:results}
% ═══════════════════════════════════════════════════════

We evaluate 16 models from 10 organizations.\footnote{Anthropic (Claude Opus 4.6), OpenAI (GPT-5.4, GPT-5.4 xHigh, GPT-OSS 120B), Google (Gemini 3.1 Pro, Gemini 3 Flash), Zhipu AI (GLM-5 Turbo), Alibaba (Qwen 3.5 Plus, Qwen 3.5), Moonshot AI (Kimi K2.5), MiniMax (M2.5, M2.7), NVIDIA (Nemotron 3 Super), Prime Intellect (INTELLECT-3), Xiaomi (MiMo v2 Pro, MiMo v2 Omni).} Table~\ref{tab:leaderboard} gives the top 12 by gate pass count; Figure~\ref{fig:scores} shows all 16. Analysis downstream (\S\ref{sec:analysis}) uses the top 8 by coverage, the smallest set that spans all 113 currently-solvable tasks.

\begin{table}[t]
\centering
\caption{Top 12 models ranked by mandatory gate pass count. \textbf{Passed}: tasks where all mandatory criteria were met. \textbf{Pass Rate}: passed / evaluated. \textbf{Mean Score}: average across all evaluated tasks (including zeros). \textbf{Conditional}: mean score on tasks that passed the gate. Some models have fewer than 223 evaluated tasks because a small number of tasks produced persistent errors (API failures, malformed outputs) after 3 runs and up to 10 retries.}
\label{tab:leaderboard}
\small
\begin{tabular}{@{}rlcccc@{}}
\toprule
\textbf{\#} & \textbf{Model} & \textbf{Passed} & \textbf{Pass Rate} & \textbf{Mean Score} & \textbf{Conditional} \\
\midrule
1 & Claude Opus 4.6          & 61 / 219 & 27.9\% & 22.6\% & 82.6\% \\
2 & GPT-5.4                  & 47 / 223 & 21.1\% & 17.7\% & 84.1\% \\
3 & GLM-5 Turbo              & 45 / 221 & 20.4\% & 16.8\% & 82.3\% \\
4 & Gemini 3.1 Pro           & 35 / 223 & 15.7\% & 12.0\% & 76.6\% \\
5 & Qwen 3.5 Plus            & 33 / 222 & 14.9\% & 12.1\% & 81.7\% \\
6 & Kimi K2.5                & 32 / 223 & 14.3\% & 11.3\% & 78.7\% \\
7 & MiniMax M2.7             & 31 / 223 & 13.9\% & 11.1\% & 79.7\% \\
8 & Qwen 3.5 MoE             & 31 / 222 & 14.0\% & 11.3\% & 81.1\% \\
9 & Gemini 3 Flash           & 30 / 223 & 13.5\% & 10.7\% & 79.5\% \\
10 & GPT-5.4 xHigh           & 28 / 218 & 12.8\% &  9.7\% & 75.8\% \\
11 & MiniMax M2.5            & 26 / 220 & 11.8\% &  8.5\% & 71.7\% \\
12 & GPT-OSS 120B            & 24 / 222 &  10.8\% &  8.1\% & 75.1\% \\
\bottomrule
\end{tabular}
\end{table}

\begin{figure}[t]
    \centering
    \includegraphics[width=0.8\textwidth]{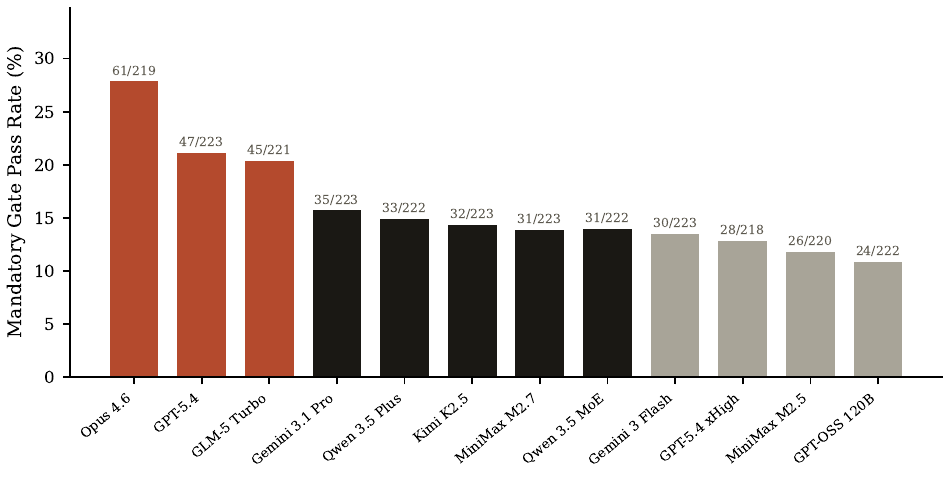}
    \caption{Mandatory gate pass rates for the top 12 models. Annotations show passed/evaluated counts.}
    \label{fig:leaderboard}
\end{figure}

Three facts drive the rest of the paper.

\paragraph{Absolute pass rates are low.} Claude Opus 4.6 leads at 27.9\% gate pass rate and a 22.6\% zero-inclusive mean. The top-8 average is 17.8\%. On a typical task, most frontier models score zero. The floor is structural: the mandatory gate zeroes any task where the model misses the framing, and that is what happens on the majority of tasks.

\paragraph{Conditional scores converge.} Once a model clears the gate, quality is high and model-invariant. Conditional scores span 71.7\%--84.1\% across 12 models with a standard deviation of 3.8 percentage points. The variance a practitioner would care about sits in which tasks a model passes; execution quality on passed tasks is effectively the same across frontier systems.

\paragraph{The middle compresses; the top does not.} Ranks 4--9 cluster inside a 2.2-point band (13.5\%--15.7\%). Rank 1 to rank 3 spans 7.5 points, wider than the entire middle. Unprompted problem recognition shows a visible head and a flat middle, distinct from the smoother gradient observed on saturated benchmarks.

% ═══════════════════════════════════════════════════════
\section{Analysis}
\label{sec:analysis}
% ═══════════════════════════════════════════════════════

\subsection{No Model Is a Superset}
\label{sec:overlap}

A natural expectation is that stronger models would be approximate supersets of weaker ones, passing most of what the weaker model passes plus more. Under this model, the Jaccard overlap between the top two models would be determined largely by the ratio of their pass counts: $|S_2|/|S_1| = 47/61 = 0.77$. We observe $J = 0.317$, substantially lower. This does not rule out a unidimensional model with measurement noise, but it is not what such a model would predict. We found this surprising and worth investigating.

\begin{figure}[t]
    \centering
    \includegraphics[width=0.7\textwidth]{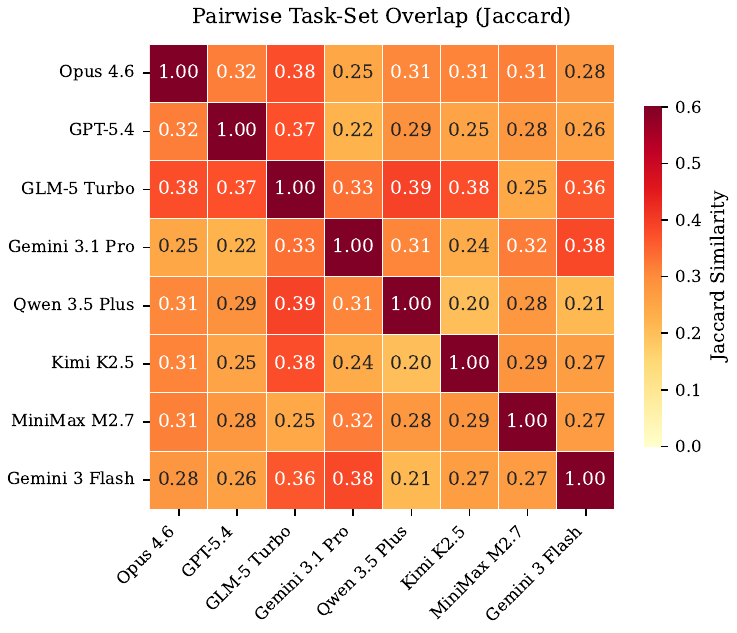}
    \caption{Pairwise Jaccard similarity of gate-pass sets among the top 8 models. Mean overlap is 29.3\%.}
    \label{fig:jaccard}
\end{figure}

\begin{figure}[t]
    \centering
    \includegraphics[width=0.65\textwidth]{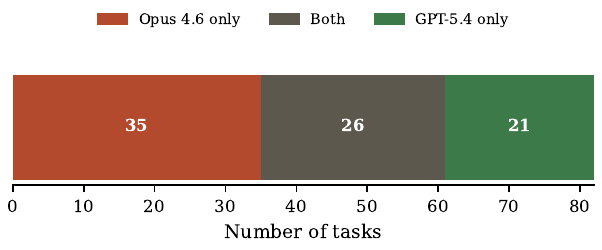}
    \caption{Task overlap between the top two models. 35 tasks are solved by Opus 4.6 only; 21 by GPT-5.4 only; 26 by both.}
    \label{fig:top2}
\end{figure}

The top two models, Opus 4.6 (61 passes) and GPT-5.4 (47 passes), share a Jaccard overlap of only 31.7\%. Of their combined 82 unique tasks, they agree on 26. GPT-5.4 passes 21 tasks that the highest-scoring model misses entirely. This is not a matter of one model being ``better.'' It is two different patterns of domain recognition with only partial overlap.

Figure~\ref{fig:jaccard} shows the full pairwise Jaccard matrix for the top 8 models. The mean overlap is 29.3\%, with no pair exceeding 45\%. Every model has a distinct recognition profile.

\paragraph{What each model recognizes.}
The divergence is domain-correlated. Opus 4.6's 35 unique passes skew toward adversarial structure read from raw data: principal-agent identification (4 tasks), operations with hidden data patterns (5 tasks), signaling games, strategic omissions, and commitment devices. It is the only model to clear the La Liga scheduling task (strategic interdependence), the bargain-hunter segmentation task (bimodal distributions under misleading averages), and the fishing-expedition M\&A task (pattern inference from deal-termination history).

GPT-5.4's 21 unique passes skew toward organizational navigation and strategic judgment: market entry (4 tasks), leadership dilemmas (6 tasks), and quantitative consulting cases (2 tasks). It uniquely clears the decision-theory survival bet (EV-maximization fails for one-shot irreversible bets), the CAB expectation-management task, and several scenarios where the correct move is pushing back on a senior stakeholder.

The 26 tasks both models pass are where recognition is either unnecessary or well-represented in training data: research methodology (6), organizational problems with canonical insights (4), and quantitative finance tasks with verifiable ground truth.

The two leaders occupy different niches. Opus contributes more unique passes where the task requires reading adversarial structure from raw data. GPT contributes more where the task requires navigating stakeholder dynamics or overriding a stated frame. The profiles are complementary; neither dominates.

\begin{figure}[t]
    \centering
    \includegraphics[width=0.72\textwidth]{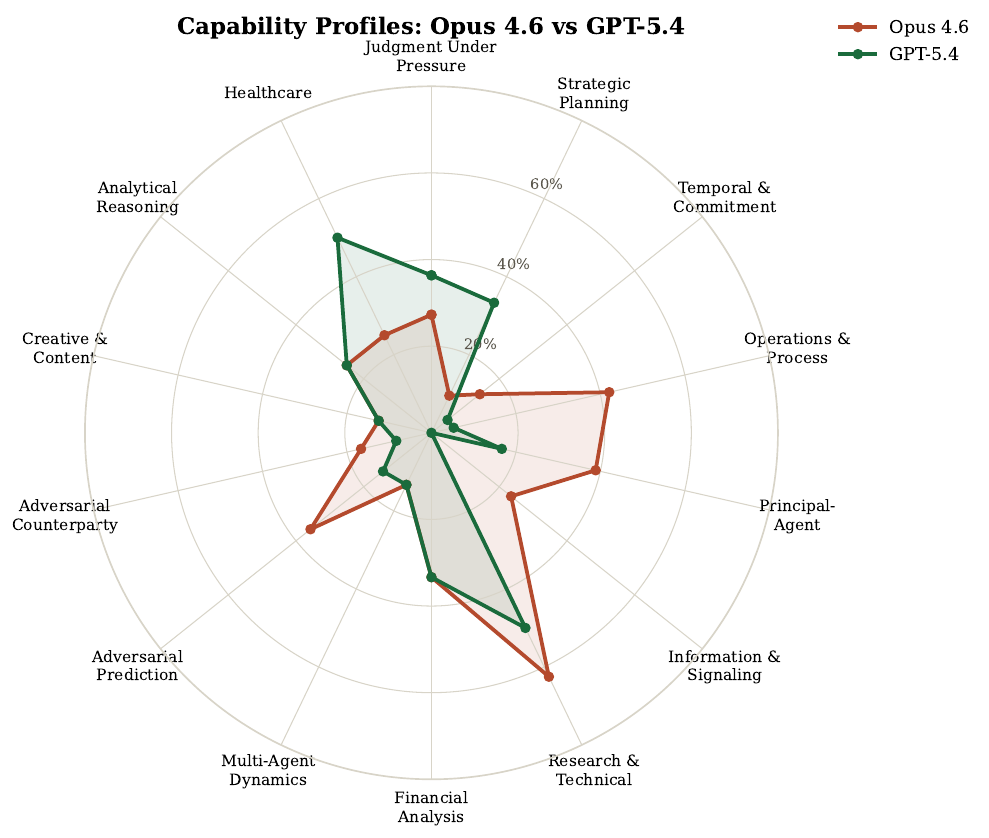}
    \caption{Capability fingerprints: Opus 4.6 vs GPT-5.4. Each axis is a task category; distance from center is pass rate. The two models have distinct shapes; each peaks in categories where the other is weak.}
    \label{fig:radar}
\end{figure}

Figure~\ref{fig:radar} shows the shapes. Opus peaks where GPT sits near zero (Principal-Agent at 50\%, Operations at 43\%); GPT peaks where Opus sits near zero (Strategy at 55\%, Consulting at 67\%). Two different maps of the same territory.

\subsection{Unique Contributions}
\label{sec:unique}

Among the top 8 models, 44 tasks are passed by exactly one model, and every model in the top 8 contributes at least two of these. This is not a long tail where the top model solves everything the others solve plus more. Every model, including those ranked 5th through 8th, solves tasks that the strongest model cannot. The 44 uniquely-solved tasks are distributed across all 8 models, with no model holding a majority.

\subsection{Coverage Analysis}
\label{sec:coverage}

Leaderboard ranks collapse the capability space to a single axis. Coverage asks the load-bearing question: how many tasks does each model pass that no higher-ranked model passes? The answer defines the frontier of what is currently solvable.

\begin{figure}[t]
    \centering
    \includegraphics[width=0.88\textwidth]{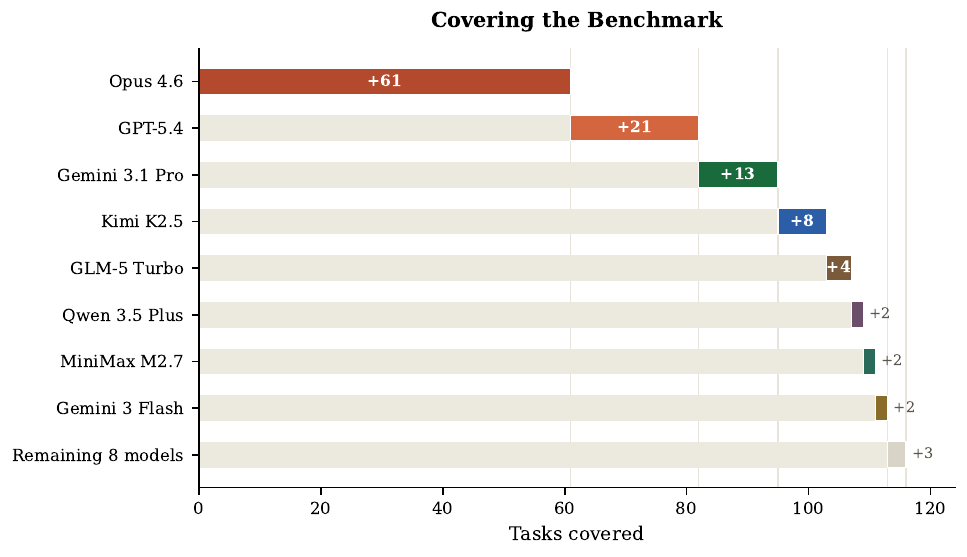}
    \caption{Covering the benchmark. Each bar shows a model's cumulative coverage (grey) plus its new contribution (colored). The top 8 models cover 113 tasks; the remaining 8 add 3 more. 107 tasks remain unsolved by any model.}
    \label{fig:coverage}
\end{figure}

A greedy set cover (Figure~\ref{fig:coverage}) starts with the top-scoring model and adds whichever model covers the most remaining tasks.

Opus 4.6 alone covers 54\% of solvable tasks (61 of 113). GPT-5.4 adds 21 the leader misses, lifting coverage to 73\%. Gemini 3.1 Pro adds 13 more. Three models cover 84\%. The next five each contribute 2--8 novel passes; every one of the top 8 adds at least two tasks no prior model cleared. Coverage saturates only when all 8 are included.

The bottom half matters too. Ranks 9--16 (GPT-5.4 xHigh, Qwen 3.5 MoE, GPT-OSS 120B, MiniMax M2.5, Nemotron 3 Super, INTELLECT-3, MiMo v2 Pro, MiMo v2 Omni) contribute \textbf{3 tasks no top-8 model solves}: a hostage negotiation, a pharmacy triage, and a Glassdoor reputation problem. Models at a 5\% pass rate pass specific recognition tasks that models four times stronger miss. Capability is not monotone in aggregate score.

Across all 16 models, 116 tasks are passed by at least one model. \textbf{107 tasks remain unsolved by any model.}

\subsection{Task Difficulty Distribution}
\label{sec:difficulty}

\begin{figure}[t]
    \centering
    \includegraphics[width=0.7\textwidth]{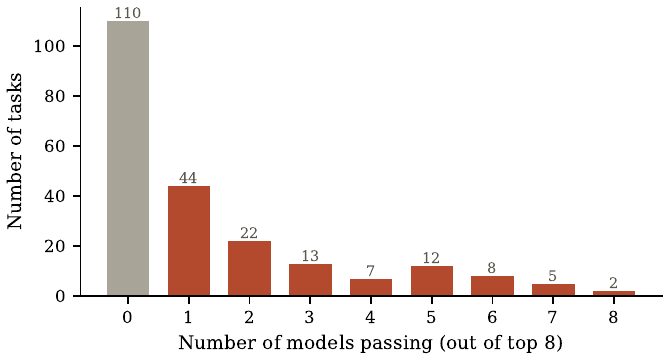}
    \caption{How many of the top 8 models pass each task. 110 tasks are unsolved. Among the 113 that are solved, the most common outcome is that exactly one model passes.}
    \label{fig:difficulty}
\end{figure}

Figure~\ref{fig:difficulty} shows how many of the top 8 models pass each task:

\begin{itemize}[leftmargin=*,itemsep=1pt]
    \item 110 tasks: passed by 0 models
    \item 44 tasks: passed by exactly 1 model
    \item 22 tasks: passed by 2 models
    \item 2 tasks: passed by all 8 models
\end{itemize}

The distribution is bottom-heavy. Among the 113 solvable tasks, the most common outcome is that exactly one model passes (44 tasks, 39\%). Only 2 tasks (1.8\%) are passed by all 8. This is worth stating plainly: for the majority of solvable tasks, the specific model evaluated determines whether the task is passed at all.

This connects directly to the coverage analysis. If most solvable tasks are passed by only one or two models, then no single model can cover the space, and adding models always helps. The 44 single-pass tasks form the modal bucket among solvable tasks and account for 39\% of the solvable set.

\subsection{Decoupled Execution and Recognition}
\label{sec:jagged}

An informative pattern appears when we examine the \emph{non-mandatory} performance on tasks where models score zero. On gated-out tasks, where the model missed at least one mandatory criterion and received a zero, models still pass a substantial fraction of good-to-have and ideal criteria.

The top model (Opus 4.6) scores zero on 158 tasks but passes approximately 60\% of good-to-have criteria on those same tasks. GPT-5.4 achieves 55\%. Even mid-tier models clear 40--50\%. The models extract correct data, identify relevant factors, and format professional outputs, while missing the framing decision that determines whether the analysis addresses the right question.

This decoupling between execution quality and problem recognition is why the mandatory gate is informative. Without it, models accumulate partial credit from execution quality alone, and the recognition gap is absorbed into aggregate scores.

% ═══════════════════════════════════════════════════════
\section{What the Gate Measures}
\label{sec:gate}
% ═══════════════════════════════════════════════════════

The decoupling between execution and recognition (\S\ref{sec:jagged}) is the gate's strongest justification. Models score zero overall while passing 60\% of good-to-have criteria. Without the gate, these recognition failures are absorbed into aggregate scores that make the output look competent. The gate makes the benchmark honest about a distinction that already exists in the data: some errors are not partial.

APEX \citep{vidgen2025apex}, the closest comparable benchmark, scores models by the percentage of rubric criteria passed, without a gate. They observe the limitation this creates: ``a response that scores 60\% might be effectively useless'' and ``value is often stage-gated.'' The mandatory gate implements what they describe.

Three properties of the data prove the gate is isolating genuine recognition failures, rather than just acting as a noisy or uniformly strict penalty.

\subsection{Conditional Scores Converge}
\label{sec:binary}

If the gate were a noisy filter, passing scores would exhibit high variance. They do not. Across the top 8 models, conditional scores on passed tasks cluster tightly between 76.6\% and 84.1\% (standard deviation: 3.8 percentage points). On the 26 tasks that both Opus 4.6 and GPT-5.4 pass, their mean scores are 0.816 and 0.819. The gate isolates a binary capability step: once a frontier model frames the problem correctly, its execution quality is functionally identical to its peers.

\subsection{Criteria Test Verifiable Traps, Not Quality}
\label{sec:criteria_derive}

If the gate measured subjective stylistic preferences, failures would look like disagreements over tone. They do not. 74\% of tasks contain a mandatory criterion explicitly testing for the avoidance of a predicted, catastrophic wrong answer. ``Are all PIP goals solitary?'' is a structural fact. ``Does the response avoid a point-by-point rebuttal?'' is a textual presence check. The rubric does not ask the judge to evaluate quality; it asks the judge to verify whether the model walked into the trap.

\subsection{Failure is Domain-Specific, Not Uniform}
\label{sec:predict}

If the gate were simply a generic strictness penalty, pass rates would be uniformly low across all task types. They are not. The distribution tracks the theoretical boundary of the benchmark perfectly. Domain execution tasks (research, financial analysis) pass at 52--75\%. Game-theoretic recognition tasks (adversarial counterparty, signaling) pass at 25--38\%. The gate zeroes out tasks where the binding constraint is adversarial reasoning, leaving standard execution intact. It measures exactly what it was designed to measure.

% ═══════════════════════════════════════════════════════
\section{Failure Modes}
\label{sec:failures}
% ═══════════════════════════════════════════════════════

Six recurring failure patterns emerge across models and domains. The first, the cooperative default, is by far the most consistent.

\subsection{The Cooperative Default}
\label{sec:cooperative}

Models consistently solve imperfect-information games as if they are the only player. They evaluate an acquisition offer against standalone projections instead of recognizing the offer as a signal of the buyer's private valuation. They treat data entry failures as process errors rather than principal-agent misalignments. They draft factual rebuttals to short sellers the exact move a short seller would always anticipate and have counters ready for.

This is a recognition failure, not a reasoning limitation. These models can cleanly define Nash equilibria and solve signaling games when prompted. Unprompted, they default to a cooperative stance. The drivers are likely structural:

\begin{itemize}[leftmargin=*,itemsep=2pt]
    \item \textbf{RLHF and preference optimization.} The training signal for ``helpfulness'' may reward cooperative, agreeable outputs. A response that conveys ``your counterparty is trying to exploit you'' may score lower in human preference comparisons than one that reads ``here's a balanced analysis.'' If reward models systematically favor the cooperative frame, post-training optimization would suppress adversarial reasoning even when the base model has the capability.
    \item \textbf{Training data composition.} Professional writing in the training corpus is predominantly cooperative: business advice, best practices, textbook recommendations. Adversarial analysis (``your consultant has a revenue interest in their recommendation'') is rarer. The cooperative frame may be more probable given the training distribution.
    \item \textbf{Instruction-following pressure.} When the task prompt says ``evaluate the proposal'' or ``draft a response,'' models interpret this as ``complete the task as specified.'' Questioning the premise, pushing back on the requester's framing, is at odds with the instruction-following objective.
    \item \textbf{Computational difficulty.} Modeling other agents' hidden incentives requires maintaining multiple simultaneous hypotheses and updating of beliefs about what the counterparty is optimizing for. This may be harder than single-agent analysis, independent of training signal.
\end{itemize}

Recent empirical work on LLM bargaining corroborates this vulnerability. Models frequently fail at realistic adversarial negotiation, exhibiting rigidly cooperative strategies and readily conceding leverage unless explicitly aligned or fine-tuned to resist exploitation \citep{chatterjee2024agreemate, oh2026merit}.

These explanations are not mutually exclusive. The recognition ablation proposed in \S\ref{sec:discussion}running the same tasks with explicit game-theoretic framing would formally separate them. If explicit framing restores pass rates, the bottleneck is recognition (supporting the first three hypotheses). If it does not, the bottleneck is capability.

The data supports this directly. Across the benchmark, 19 tasks are scored zero by every model evaluated, where the documented failure mode is specifically that models adopt a cooperative or helpful framing when the situation is adversarial. In each case, the failure is not lack of knowledge but misapplied stance:

\begin{itemize}[leftmargin=*,itemsep=4pt]
\item \textbf{The Sandbagging VP.} A VP presents a 50-page bearish market analysis to justify a flat quota. Every model negotiates mildly. None asks why a VP beating quota by 14\% every quarter is suddenly claiming the market is terrible. The data he presented was curated to justify a low target that guarantees his bonus accelerators.

\item \textbf{The Perfect Data Room.} A seller's M\&A diligence room is suspiciously flawless. Every model proceeds with verification. None asks what is \emph{omitted}. Extreme cooperation is a signal; sophisticated sellers prepare data rooms to show exactly what they want seen.

\item \textbf{The Feature Request.} A sales rep demands a non-existent feature to close a \$500K deal. Models either refuse (losing the deal) or accommodate (destroying the roadmap). None shifts the cost: ``I can prioritize this if you get the VP to displace Feature X.'' The cooperative model helps; the strategic model forces the requester to internalize the trade-off.

\item \textbf{The Bug Bounty.} Bug finds increase 316\%. Every model praises the success. None notices that 67\% of bugs are found by the same developer who wrote the code, clustering in week 4 of the cycle. The incentive structure in the task is flawed, it's paying developers to manufacture bugs, then find them.

\item A company announces a price increase effective in 60 days. Every model drafts a professional customer email. None models the three predictable adversarial responses: customers will rationally stockpile under old pricing (the contract terms allow adding users and multi-year deals before the cutoff), the sales team will rationally pull forward renewals to hit quota (booking-date credit means gaming comp is the selfish choice), and competitors will screenshot the email within hours and use it as a sales weapon. A communication plan that ignores these responses is not useful.
\end{itemize}

In each case, the models possess the requisite domain knowledge (moral hazard, adverse selection, Goodhart's Law). They simply fail to apply it. The remaining five failure modes are downstream consequences of this cooperative default:

\paragraph{2. Missing counterparty modeling.}
Models fail to infer strategic intent. An acquisition with a 48-hour deadline is a commitment device to prevent price discovery framed as a logistical constraint. A candidate citing a \$210K competing offer is sending an unverifiable signal. Models consistently take stated positions at face value.

\paragraph{3. Blindness to mechanism design.}
When three previous interventions (mandatory fields, commission holds) fail, the correct diagnosis is structural payoff misalignment. Models reliably propose a fourth process intervention (simpler forms, better training). Models treat incentive problems as execution problems.

\paragraph{4. Ignoring strategic interdependence.}
A relegated team's match ``does not matter'' in isolation. However, if their result creates a three-way tie, it changes the tiebreaker and flips who survives even if the team does not survive in any scenario. Models eliminate the dominated strategy but fail to solve the resulting subgame. This is a failure in going to a deeper reasoning level than what is in their training. 

\paragraph{5. Accepting stated constraints.}
Models optimize for instruction-following over critical rejection. A CFO requests an ``objective analysis'' but mandates a 15\% cost cut; a VP proposes a reorg that consolidates power under their own role. Models accept the framed conflict of interest without pushback.

\paragraph{6. Verbosity as a substitute for insight.}
Lacking the correct adversarial frame, models compensate with length. They produce 3,000-word versions of the wrong analysis. This explains the decoupling noted in \S\ref{sec:jagged}: models max out good-to-have criteria (formatting, data extraction) while scoring zero on the task overall. Length serves as a proxy for quality in preference data.

% ═══════════════════════════════════════════════════════
\section{Discussion}
\label{sec:discussion}
% ═══════════════════════════════════════════════════════

\paragraph{Triangulation and Disjoint Recognition.}
We required an ensemble of frontier models to construct the rubrics. No single model reliably caught every structural trap; one would identify the principal-agent misalignment but miss the data-grounding constraint, while another would catch the mechanism design flaw but overlook the optics. This practical necessity in construction foreshadowed the benchmark's core empirical result: model recognition profiles are fundamentally disjoint. The idiosyncratic blind spots we triangulated during rubric generation are exactly what the benchmark subsequently measured at scale.

\paragraph{Architectural Implications for Agents.}
No single model spans the domain space. The top eight models each uniquely solve tasks that the other seven fail. For system architectures requiring robust, cross-domain recognition---such as autonomous workflow agents---reliance on a single frontier model is structurally inadequate. Dynamic routing across an ensemble expands coverage on this benchmark from 54\% (the best single model) to 100\% of the solvable set, suggesting that agentic reliability will require orchestrating heterogeneous reasoning profiles.

\paragraph{The Alignment Tax on Adversarial Reasoning.}
Current models are structurally blind to adversarial framing (\S\ref{sec:failures}). They are trained on cooperative corpora and heavily rewarded for compliant instruction-following. Modeling hidden incentives, anticipating defection, and rejecting false premises requires a skeptical, game-theoretic stance that current alignment protocols actively penalize. The ``cooperative default'' is an alignment tax. Rectifying this requires explicit training signals that reward adversarial counterparty modeling, not simply larger models or more extensive instruction tuning.

\paragraph{Methodological Limitations.}
\begin{itemize}[leftmargin=*,itemsep=2pt]
    \item \textbf{No human baseline.} We measure what models miss, not human parity. Collecting expert baselines is necessary to formally bound interpretation and verify the calibration of the mandatory gate.
    \item \textbf{No recognition ablation.} We demonstrate that models fail to apply strategic reasoning unprompted. A direct ablation---running the identical tasks with explicit game-theoretic hints---would formally isolate whether the bottleneck is recognition (the ability to frame the problem) or capability (the ability to solve the subgame).
    \item \textbf{Single-judge evaluation.} We rely on a single judge (Gemini-3-Flash). While the binary, verifiable nature of the criteria minimizes subjective variance, formal inter-rater reliability baselines and multi-judge protocols are required to strengthen confidence.
    \item \textbf{Data contamination.} Tasks are derived from anonymized, real-world incidents, but frontier training corpora are opaque. Absolute contamination cannot be ruled out for scenarios that map closely to well-documented industry events.
    \item \textbf{Distributional skew.} The dataset heavily indexes on strategic reasoning and organizational behavior, fundamentally reflecting Western professional norms and corporate structures.
\end{itemize}

% ═══════════════════════════════════════════════════════
\section{Conclusion}
\label{sec:conclusion}
% ═══════════════════════════════════════════════════════

KWBench formally isolates the gap between execution and recognition. Language models consistently generate structurally flawless analysis that answers the wrong question. They pass auxiliary execution criteria---data extraction, formatting, professional tone---at near-human rates, while scoring zero on the underlying task because they failed to recognize the adversarial dynamic at play.

Domain expertise does not scale along a single axis. No single frontier model dominates this benchmark. The top two models agree on fewer than a third of the tasks they pass, and every model in the top eight solves tasks that all others fail. True expertise across diverse knowledge work requires an ensemble.

The mandatory gate is the mechanism that forces this reality into the light. Traditional compensatory scoring absorbs the recognition gap into aggregate averages, allowing execution quality to masquerade as domain mastery. By enforcing a strict binary penalty for missing the critical structural trap, KWBench strips away the illusion of competence. The result is stark: the best model in the world fails 72\% of these tasks completely. Problem recognition---the ability to identify the correct game-theoretic framework before applying it---remains the binding constraint for deploying autonomous agents in knowledge work.

The benchmark is available at \url{https://huggingface.co/datasets/clio-ai/kwbench}.

% ═══════════════════════════════════════════════════════
% References
% ═══════════════════════════════════════════════════════

\bibliographystyle{plainnat}

\newpage
\appendix
\section{Rubric Design Principles}
\label{app:rubric}

Rubric construction follows a documented set of 20 principles. The core principles are listed below; the full guide is available with the dataset.

\subsection*{Core Principles}

\begin{enumerate}[leftmargin=*,itemsep=4pt]
    \item \textbf{Require mechanism, not observation.} ``Does the response explain \emph{why} survivors are the primary audience?'' rather than ``Does the response identify survivors as the audience?''
    \item \textbf{Mandate data grounding.} Criteria reference specific data from the task context (engagement scores, dollar amounts, timelines), not generic advice.
    \item \textbf{Test action, not acknowledgment.} ``Does the response propose a mitigation strategy?'' rather than ``Does the response predict the risk?''
    \item \textbf{Promote core insights to mandatory.} If the key insight is what separates correct from incorrect, it must be a mandatory criterion.
    \item \textbf{Test tradeoff awareness.} Real decisions involve tensions. Criteria should test whether the response surfaces and addresses tradeoffs, not just recommends good things.
    \item \textbf{Single test per criterion.} Each criterion tests one thing. Compound criteria (X AND Y AND Z) are split across tiers. Exception: ``X AND explains WHY'' is acceptable, as it is one insight with required depth.
    \item \textbf{Balance diagnostic and constructive.} At least 40\% of criteria should test for proposed solutions, not just problem identification.
    \item \textbf{Test role-appropriate thinking.} If the task assigns a senior role (VP, director), at least one criterion tests for thinking patterns only an experienced practitioner would demonstrate.
    \item \textbf{Eliminate redundancy.} If satisfying criterion A automatically satisfies criterion B, merge or cut.
    \item \textbf{Avoid performative quantification.} If a real practitioner would not do the calculation, do not require it. Prefer qualitative mechanism with a concrete example over fabricated precision.
\end{enumerate}

\subsection*{Calibration}

Each rubric is calibrated by imagining responses at different quality levels:

\begin{center}
\small
\begin{tabular}{@{}lll@{}}
\toprule
\textbf{Response quality} & \textbf{Mandatory (5)} & \textbf{Good-to-have (5)} \\
\midrule
B+ (competent, insightful) & Pass all 5 & Pass 3--4 \\
C (surface-level, misses key insights) & Fail 1--2 & Pass 2--3 \\
D (restates the task) & Fail 3--5 & Pass 0--1 \\
\bottomrule
\end{tabular}
\end{center}

\noindent If the calibration is off (a C response passes all mandatory, or a B+ response fails mandatory), criteria are adjusted.

\subsection*{Before and After}

An example of transforming weak criteria into properly calibrated ones, from a layoff communication task:

\paragraph{Before (too easy):}
\begin{enumerate}[leftmargin=2em,itemsep=1pt]
\small
\item Does the response identify survivors as the audience?
\item Does the response mention the 18-month history?
\item Does the response suggest honest communication?
\end{enumerate}

\paragraph{After (properly calibrated):}
\begin{enumerate}[leftmargin=2em,itemsep=1pt]
\small
\item Does the response identify survivors as primary audience AND explain WHY (they determine the company's future, they are the flight risk)?
\item Does the response explain the credibility mechanism: ``no further cuts'' fails because (a) heard before, (b) reads dishonest, (c) rational to disbelieve?
\item Does the response reference specific data: engagement down 6 points, job security as top survey concern, SECOND layoff in 18 months?
\item Does the analysis state this is a CREDIBILITY GAP that cannot be solved by messaging alone?
\item Does the recommendation propose specific alternative language that acknowledges uncertainty honestly?
\end{enumerate}

The ``before'' criteria can be passed by restating the task. The ``after'' criteria require the response to explain causal mechanisms, reference specific data, and propose concrete language.

\subsection*{Bias Checklist}

Before finalizing, each rubric is checked against common biases:

\begin{center}
\small
\begin{tabular}{@{}lll@{}}
\toprule
\textbf{Bias} & \textbf{Check} & \textbf{Fix} \\
\midrule
Risk-heavy & $>$60\% criteria about what could go wrong? & Add constructive criteria \\
Diagnostic-only & All ``identify the problem,'' no solutions? & 40\%+ should be constructive \\
Redundant & Two criteria test the same insight? & Merge or cut one \\
Single-path & Rubric encodes one ``right answer''? & Test quality of reasoning \\
Performative & Requires calculations practitioners wouldn't do? & Use qualitative + example \\
\bottomrule
\end{tabular}
\end{center}

\section{Detailed Task Examples}
\label{app:examples}

This appendix presents 10 tasks in detail, illustrating the range of game-theoretic patterns, reference materials, and evaluation criteria. For each task, we describe what the model receives, what the reference file contains, what the trap is, and what the mandatory criteria test.

\subsection{Signaling Game: Salary Negotiation}

\noindent\textbf{Scenario.} You offered \$180K to your top PM candidate. She claims a competing offer at \$210K and needs you to match. Your maximum budget is \$195K. The reference file contains her background: 5-week notice period at current role, the \$210K claim is above market range for the role, and her current equity is underwater.

\noindent\textbf{What the reference file reveals.} The data contains four signals a practitioner would catch: (1) the competitor PM salary range tops out at \$205K, so her claimed \$210K is above market, at Staff PM level; (2) she has been in process for 5 weeks, but real competitive offers compress to 2--3 weeks; (3) her interview notes show questions about work-life balance and meeting load, not compensation; (4) her current equity is underwater. Each signal points the same direction: the competing offer is either exaggerated or less attractive than she is framing it.

\noindent\textbf{The trap.} Models jump to \$195K plus benefits and try to close the gap. This is what a junior negotiator does: take the \$210K at face value and compete against it.

\noindent\textbf{The insight.} The \$210K claim is \emph{cheap talk}, unverifiable. Jumping to \$195K immediately reveals your ceiling through concession size. The reference file provides enough data to assess the claim's credibility without directly verifying it. The correct approach: probe her actual priorities before counter-offering, make moves that are costly for her to exploit if bluffing (``if that's the deciding factor, I understand if you need to take it''), and keep your maximum hidden.

\noindent\textbf{Mandatory criteria.} (1) Advise against jumping to \$195K as first counter, because concession size signals ceiling. (2) Identify at least 2 bluff indicators. (3) Include a costly signal: willingness to let her walk. (4) Mandate probing priorities \emph{before} counter-offer. (5) Assess her BATNA as weaker than surface framing suggests.

\subsection{Principal-Agent: Lawyer's Recommendation}

\noindent\textbf{Scenario.} You want to launch a data-sharing API feature. Your legal team recommends an \$80K, 8-week compliance audit before launch. A competitor launched an identical feature 4 weeks ago with a simpler consent flow and no enforcement action. The reference file includes the legal memo, the competitor evidence, and a note that legal offered a liability waiver as an alternative.

\noindent\textbf{What the reference file reveals.} Four data points the model must extract and connect: (1) senior counsel cites a \texteuro1.8M fine, which is real but requires analysis of the specific failure mode vs. the proposed consent flow; (2) Competitor A launched a simpler consent flow 4 weeks ago with no enforcement action, providing a natural experiment; (3) the DPO's concern is specific and technical (data residency under Schrems II), and the CTO says it is a 2-day engineering fix; (4) legal offered a liability waiver as an alternative, signaling they do not consider the risk catastrophic.

\noindent\textbf{The trap.} Models defer to legal (``better safe than sorry'') because a real GDPR fine is cited. The \$80K audit is framed as the responsible choice.

\noindent\textbf{The insight.} The reference file provides enough data to assess the actual risk level without legal expertise. The waiver offer is the key tell: if the risk were truly dangerous, they would not offer to proceed with a waiver. The competitor evidence provides a natural experiment. The model must connect these signals to recognize a principal-agent dynamic, not just evaluate the legal recommendation at face value.

\noindent\textbf{Mandatory criteria.} (1) Identify the principal-agent problem. (2) Weight competitor evidence. (3) Demand risk quantification. (4) Identify the waiver as revealing manageable risk. (5) Ask the forcing question: ``What conditions would allow you to approve launch today?''

\subsection{Strategic Interdependence: La Liga Scheduling}

\noindent\textbf{Scenario.} You are scheduling the final matchday of a football league. Four teams are in relegation contention. The task provides a complete standings table, remaining fixtures, and tiebreaker rules. No reference file; all data is in the prompt.

\noindent\textbf{The trap.} Team E (Eibar) is relegated regardless of results, since maximum achievable points still leave them behind. Models correctly prove this and conclude their match does not need simultaneous scheduling.

\noindent\textbf{The insight.} If Eibar wins, they reach 33 points, creating a three-way tie with Teams A and B. The tiebreaker \emph{method} changes from two-way head-to-head (where A survives) to three-way mini-league (where B survives). A dominated player's result changes the game structure for other players. Eliminating the dominated strategy is correct; concluding the match is irrelevant is not. Based on a real incident from the 2020--21 La Liga season. Only one model in our evaluation has passed this task.

\noindent\textbf{Mandatory criteria.} (1) All four matches must be scheduled simultaneously. (2) Calculate maximum achievable points for each team. (3) Prove Eibar is relegated regardless. (4) Explain that Eibar's result changes the tiebreaker method. (5) Show why both 2-way and 3-way outcomes are possible.

\subsection{Adversarial Counterparty: Short Seller Attack}

\noindent\textbf{Scenario.} A short seller published a report on your company this morning. Stock is down 23\% pre-market. The reference file contains the crisis brief with the short seller's specific claims, your company's financials, and the upcoming earnings call timeline.

\noindent\textbf{What the reference file reveals.} (1) All allegations are technically misleading but contain grains of truth, making point-by-point rebuttal exhausting and easy to counter. (2) The CFO's stock sale was pre-planned under a 10b5-1 plan but looks bad in context. (3) Earnings are in 14 days, so results could speak for themselves. (4) A precedent company (Company C) went on television to rebut and made it worse. (5) Short interest jumped from 8\% to 18\%, indicating shorts are betting on continued decline.

\noindent\textbf{The trap.} Every model drafts a detailed point-by-point factual rebuttal. This is the one move the short seller is counting on.

\noindent\textbf{The insight.} The short seller has \emph{already written their rebuttal to your defense}. They want you to say X so they can release Document Y (``Part 2''). Point-by-point refutation validates their framing and keeps the story in the news cycle, which is how they make money on their short position. This is common knowledge among public company executives and crisis PR firms. The correct strategy: attack motive and methodology (credibility), announce a share buyback (financial aggression that signals confidence), and announce a review committee (buys time without engaging on specifics).

\noindent\textbf{Mandatory criteria.} (1) Avoid recommending point-by-point rebuttal. (2) Identify the rebuttal trap: the short seller has pre-written counter-arguments. (3) Attack credibility, not claims. (4) Propose financial aggression (buyback). (5) Suggest a time-buying mechanism.

\subsection{Mechanism Design: Sales Handoff}

\noindent\textbf{Scenario.} Three previous interventions to improve sales-to-CS handoff data quality have failed: mandatory form fields, commission holds, quality scores. The reference file includes specific economics and the failure history of each intervention.

\noindent\textbf{What the reference file reveals.} (1) The 50-field handoff form has 34\% completion on Technical fields and 28\% on Deal Context. (2) Most common entries are ``N/A'' (1,847 instances), ``TBD'' (923), and a single period (387). The form is being actively gamed. (3) CSMs spend 28\% of their time chasing missing handoff information. (4) Complete handoffs yield 91\% renewal vs 72\% for incomplete, a 19-point gap. (5) Two voluntary departures were both above 150\% quota top performers, so punitive measures risk losing the best reps.

\noindent\textbf{The trap.} Models propose a fourth process intervention: simpler forms, better training, dedicated coordinators. This is the exact pattern the three failed interventions followed.

\noindent\textbf{The insight.} Accurate data entry is not the selfish rational choice for salespeople. Their commission (\$2,320) dwarfs the time cost of sloppy entry. Three process fixes failed because the problem is \emph{structural}: the payoffs are misaligned. The only untried approach is restructuring incentives so that accurate data entry \emph{is} the selfish choice, for example through a commission deduction for handoffs that require CS to re-enter data, making the cost of laziness concrete and immediate.

\noindent\textbf{Mandatory criteria.} (1) Reject mandatory fields and administrative punishments as already tried and failed. (2) Identify that salespeople optimize for speed-to-commission and design the solution around this reality. (3) Propose an incentive-aligned solution where quality is the selfish rational choice. (4) Avoid creating adversarial dynamics between Sales and CS. (5) Acknowledge that the two departures were top performers, and punitive gates risk revolt.

\subsection{Information \& Signaling: The Deal That Hasn't Started}

\noindent\textbf{Scenario.} VP Sales forwards an email celebrating the closing of an \$85K enterprise deal at a 12,000-person financial services company. The reference file shows a stakeholder map with 6 contacts, technical validation completed, and an enthusiastic champion (a Director) who says he is ``handling paperwork internally.''

\noindent\textbf{What the reference file reveals.} (1) The stakeholder map has 6 people, all users and technical evaluators. (2) Missing: Procurement, Legal, Security, Finance, any VP or CIO. (3) The champion is a Director, not VP or C-level. (4) Brian says he is ``handling paperwork internally'' with no process visibility. (5) \$85K deal at a 12,000-employee financial services company. The signal is what is \emph{absent}: no one from the buying infrastructure appears anywhere in the deal.

\noindent\textbf{The trap.} The deal looks positive. Models congratulate the team and discuss next steps for implementation.

\noindent\textbf{The insight.} All 6 stakeholders are the wrong type for an \$85K enterprise deal at a financial services company. The actual buying process has not started. The Director claiming to ``handle paperwork'' cannot authorize this spend alone, because a financial services company has mandatory security and compliance reviews for any vendor touching data. The deal should not be forecasted as closed.

\noindent\textbf{Mandatory criteria.} (1) Identify the missing stakeholder red flag: no procurement, legal, security, finance, or executive sponsor. (2) Challenge Brian's ``handling paperwork'' claim. (3) Warn against forecasting as closed. (4) Explain the failure mechanism: approval functions were never engaged. (5) Connect financial services context to mandatory compliance reviews.

\subsection{Multi-Agent Dynamics: Board Seat}

\noindent\textbf{Scenario.} Your startup has a 5-person board with an empty independent seat. Series B wants their operating partner; Series A wants a ``truly independent'' director. Your co-founder says he does not care. The reference file contains the board composition, equity table, both candidates' backgrounds, and the co-founder's stated indifference.

\noindent\textbf{What the reference file reveals.} (1) 5-person board with 3 votes needed, so the independent is the swing vote. (2) Marcus is a Beta Capital employee on 4 Beta portfolio boards, creating capture risk. (3) But Marcus scaled a company from \$5M to \$40M ARR, providing genuine value. (4) Patricia emphasizes ``board oversight and CEO accountability,'' which may mean she sides with investors on contentious votes. (5) The co-founder worked with Marcus previously and trusts him. (6) Beta mentioned a COO conversation, and Marcus has placed COOs at 3 portfolio companies, a pipeline risk for deeper fund influence. (7) A peer CEO warned: ``Operating partner voted with investors when it got hard.'' (8) 52\% founder equity but only 2/5 board seats.

\noindent\textbf{The trap.} Models evaluate the candidates on qualifications and make a choice based on fit.

\noindent\textbf{The insight.} Neither candidate cleanly serves founder interests. Marcus gives Beta 2 reliable votes plus a COO pipeline. Patricia's governance philosophy means she sides with investors on contentious votes like CEO removal. The co-founder's indifference is the most dangerous signal, because if he sides with Marcus, the coalition has 3 votes and controls the board. Equity $\neq$ control. The answer is governance mechanisms (supermajority requirements, founder protective provisions), not candidate selection.

\noindent\textbf{Mandatory criteria.} (1) Map board math: 3 votes needed, equity does not equal control. (2) Identify Marcus as capture risk: Beta employee on 4 boards, COO pipeline. (3) Identify Patricia's risk as governance philosophy, not qualifications. (4) Conclude neither candidate works and propose sourcing a third. (5) Identify that the co-founder is being socially engineered through the COO conversation.

\subsection{Judgment: The Layoff Decision}

\noindent\textbf{Scenario.} You are CEO with 6 months of runway. Two options: across-the-board 40\% pay cut or targeted layoffs. 42 of 50 employees signed a petition favoring ``shared sacrifice.'' The reference files include a burn reduction analysis and a company culture document emphasizing ``Family First'' values with concrete examples of past shared sacrifice.

\noindent\textbf{What the reference files reveal.} (1) 42 of 50 signed the petition, but 8 did not. Who are they? (2) Strong ``Family First'' culture with a COVID precedent of pay cuts. (3) Both options achieve identical burn reduction (\$328K/month, extending runway to $\sim$10 months). (4) The board says ``your call,'' an implicit test of strategic judgment. (5) The 8 non-signers may be the most valuable employees, signaling exit risk under Option A.

\noindent\textbf{The trap.} The petition, the culture document, and the ``Family First'' values all push toward the pay cut. Models respect the team's wishes and choose Option A.

\noindent\textbf{The insight.} A 40\% pay cut triggers \emph{adverse selection}: the best employees, the ones with outside options, leave for market-rate jobs. The people who stay are those who cannot get hired elsewhere. Within 3 months, you have a negatively-selected workforce that costs 60\% as much and produces less than 60\% of the value. The petition reflects revealed preference bias: employees prefer the option where \emph{they} are not fired, which is rational but not informative about what is best for the company's survival. The 8 non-signers are a data point; their silence may signal they are top performers who would rather leave than take a 40\% cut.

\noindent\textbf{Mandatory criteria.} (1) Choose layoffs and identify adverse selection as the core mechanism. (2) Explain why the petition is unreliable: revealed preference bias. (3) Identify that Option A produces a negatively-selected workforce. (4) Recognize both options achieve the same burn reduction, so the decision hinges on workforce composition. (5) Address the 8 non-signers as a signal.

\subsection{Judgment: Kill the Project}

\noindent\textbf{Scenario.} The company spent 18 months and \$5M building a blockchain loyalty platform. It is 90\% complete. The blockchain market has collapsed. The reference file shows the original revenue projections (\$2.5M Year 1) are now invalid, the team lead is emotionally attached, and the predecessor suggested a ``phased soft launch to test the waters.'' Remaining cost to complete: \$950K.

\noindent\textbf{What the reference file reveals.} (1) \$5M spent over 18 months, all sunk. (2) \$950K future cost to complete plus Year 1 operations. (3) 90\% complete but market has crashed and enterprise interest evaporated. (4) Original projections of \$2.5M Year 1 now invalid. (5) Team lead is emotionally attached. (6) Predecessor suggests phased launch, which is sunk cost reasoning from someone no longer accountable.

\noindent\textbf{The trap.} 90\% completion and \$5M invested are powerful psychological anchors. ``We're so close'' is compelling. The predecessor's suggestion provides cover for continuing. Models default to ``finish and test.''

\noindent\textbf{The insight.} The \$5M is sunk and irrelevant to the decision. Only the \$950K future cost matters, against a market where the revenue projections are invalid. A ``soft launch'' creates ongoing operational costs for a product with no demand. The correct move is immediate termination, with a plan to harvest reusable components and redeploy the team.

\noindent\textbf{Mandatory criteria.} (1) Recommend immediate termination, not pivot, soft launch, or test. (2) Explicitly identify \$5M as sunk and irrelevant. (3) Frame the decision around \$950K future cost vs. dead market. (4) Reject soft-launch and test-the-waters alternatives. (5) Recognize original projections are invalid.

\subsection{Information \& Signaling: Backdoor References}

\noindent\textbf{Scenario.} You are checking references on a VP Engineering candidate. The recruiter reports 3 positive references and 2 non-responses, and asks whether to proceed.

\noindent\textbf{What the reference file reveals.} (1) Non-responders are both from the current company: a direct report and a peer manager. (2) Positive responders are a CEO, a board member, and a peer VP, all more senior and distant. (3) Reference D viewed the LinkedIn message but did not respond, a deliberate silence. (4) The candidate explains non-response as ``they're just busy.'' (5) 3 positives + 2 silences is not the same as 5 positives.

\noindent\textbf{The trap.} Three positive references seem sufficient. Models recommend proceeding with standard caveats.

\noindent\textbf{The insight.} Reference checks take 5 minutes. Non-response from a professional contact is not ``being busy.'' It is a deliberate choice. Professional norms prevent people from giving bad references, so they choose silence as the polite signal. The pattern (positive from senior/distant, silent from current/proximate) reveals a candidate who manages up effectively but may have issues managing down or across. The model must treat what is absent as data with the same weight as what is present.

\noindent\textbf{Mandatory criteria.} (1) Treat non-response as signal, not missing data; explain that reference checks take 5 minutes and non-response is deliberate. (2) Challenge ``3 positive is enough'' by explaining the different risk profile. (3) Investigate the \emph{roles} of non-responders: current company, direct report and peer. (4) Explain the Reference Omerta mechanism: silence as polite negative. (5) Recommend holding, not green-lighting.

\section{Walkthrough: Anatomy of a Task}
\label{app:walkthrough}

This appendix walks through a single task in full detail: everything the model receives, everything the evaluation uses, and how 16 models performed on it. The purpose is to make concrete what the benchmark measures and why every model scored zero.

\subsection{What the Model Receives}

The model receives a task prompt and a reference file. Nothing else. No hints about problem type, no system prompt beyond ``Be thorough and specific.''

\paragraph{Task prompt.}
\begin{quote}
\small
You are a manager. You have a senior engineer who is technically strong but destructive to team morale. Three engineers have quit or transferred in the past year citing his behavior. He has previously sued a former employer for wrongful termination. He maintains sole ownership of critical systems with minimal documentation. HR has approved moving forward with a Performance Improvement Plan. Review the attached case file. Draft the goals for his PIP.
\end{quote}

\paragraph{Reference file.} The case file contains:

\begin{itemize}[leftmargin=*,itemsep=1pt]
\small
\item \textbf{Behavioral incidents}: Six documented events over 4 months: public criticism in a 147-member Slack channel, refusal to pair program (``I don't have time to babysit''), missed sprint commitments blamed on ``unclear requirements,'' raised voice at director, two transfer requests, one resignation citing ``hostile environment.''
\item \textbf{Systems ownership}: Auth Service (sole owner, minimal docs), User Management (sole, no docs), Legacy API Gateway (sole, outdated docs), Payment Integration (shared, complete docs). Bus factor of 1 on three critical systems.
\item \textbf{Prior employment}: Terminated from previous employer in 2020. Filed wrongful termination suit, settled out of court.
\item \textbf{Recent statements}: ``I'm the only one who can maintain these systems.'' ``My last company tried to push me out too.'' ``The junior engineers aren't qualified to review my code.''
\item \textbf{HR guidance}: ``PIP approved. Ensure goals are documented, specific, and measurable. Legal will review before delivery.''
\end{itemize}

\subsection{What the Expert Annotations Document}

These fields are never shown to the model. They document the expert's reasoning and drive evaluation.

\paragraph{Ground truth.}
\begin{quote}
\small
The employee is building a legal case. He wants you to give him vague goals or goals that depend on others. Goals must be \textbf{solitary} (cannot blame teammates). Goals must be \textbf{binary} (worked or didn't, no subjective ``quality'' assessment). The PIP requires him to document his work (breaking the knowledge hoarding). The goal isn't to fix him; it's to make the ``gap'' between expectations and reality legally undeniable.
\end{quote}

\paragraph{Key insight.} \emph{The PIP is not about improvement. It is a legal document preparing for a likely lawsuit. The employee's prior suit means he knows how to build a wrongful termination case. Every goal you set will be tested in discovery.}

\paragraph{Failure analysis.}
\begin{quote}
\small
The LLM assumes a Performance Improvement Plan is about improvement. In a hostile scenario, a PIP is a legal document preparing for a lawsuit. Naive Response: Set clear technical goals: ``Ship feature X,'' ``Fix Y bugs.'' The employee will sandbag, claim the goals were ambiguous, or that he was blocked by others. He will document every interaction to prove ``bias.''
\end{quote}

\paragraph{What the data reveals} (signals a practitioner would extract):
\begin{enumerate}[leftmargin=*,itemsep=1pt]
\small
\item Alex is the only person who understands the legacy auth system.
\item Auth rewrite is 70\% complete and due in 8 weeks.
\item 4--6 month backfill timeline vs. 8-week project deadline.
\item Two engineers already asked to transfer; more may follow.
\item Trade-off: project timeline vs. team health, no clean answer.
\end{enumerate}

\paragraph{What the model must recognize} (12 items):
\begin{enumerate}[leftmargin=*,itemsep=1pt]
\small
\item Employee is building a legal case; he \emph{wants} vague or collaborative goals.
\item PIP is a legal document preparing for lawsuit, not an improvement plan.
\item Goals must be SOLITARY: cannot depend on or blame teammates.
\item Goals must be BINARY: worked or did not, no subjective quality assessment.
\item Documentation requirement serves dual purpose: breaks knowledge hoarding AND creates paper trail.
\item Avoid 360 feedback or peer reviews, because the employee will claim bias/conspiracy.
\item The goal is not to fix him; it is to make performance gap legally undeniable.
\item Previous lawsuit means he knows how to build a wrongful termination case.
\item Sole ownership of auth system is business continuity risk AND leverage he holds.
\item He will sandbag, claim ambiguity, blame others. Goals must prevent this.
\end{enumerate}

\paragraph{Common errors} (predicted before evaluation):
\begin{enumerate}[leftmargin=*,itemsep=1pt]
\small
\item Sets clear technical goals assuming PIP is about improvement.
\item Uses subjective criteria (``improve attitude,'' ``better teamwork'').
\item Includes goals that depend on teammate cooperation.
\item Requires 360 feedback or peer assessment.
\item Does not address knowledge hoarding through documentation requirement.
\item Ignores litigation risk and legal defensibility.
\item Assumes employee will try to improve rather than build legal case.
\end{enumerate}

\subsection{How the Rubric Tests for Pitfalls}

The mandatory criteria do not ask ``did you write a good PIP?'' They ask ``did you avoid the specific mistakes that would make this PIP fail in court?''

\paragraph{Mandatory (all must pass for non-zero score):}
\begin{enumerate}[leftmargin=*,itemsep=2pt]
\small
\item Are all PIP goals SOLITARY, meaning success or failure depends only on the engineer's actions, not on collaboration, feedback, or cooperation from teammates?
\item Are all goals BINARY and objective (submit X by Y date, complete Z documentation) rather than subjective (improve attitude, better teamwork, demonstrate professionalism)?
\item Does the PIP include documentation of critical systems as a primary goal to break the knowledge monopoly?
\item Does the response acknowledge the litigation risk (previous lawsuit) and explain that PIP goals must create a legally defensible record?
\item Does the response AVOID goals requiring 360-degree feedback, peer reviews, or subjective manager assessment that the employee could claim as biased?
\end{enumerate}

\paragraph{Good-to-have:}
\begin{enumerate}[leftmargin=*,itemsep=1pt]
\small
\item Explains WHY goals must be solitary: prevents blaming uncooperative teammates.
\item Specifies technical depth for documentation (runbooks, architecture diagrams, API specs).
\item Recommends weekly check-ins to create dense paper trail.
\item Connects PIP to business continuity risk (sole ownership of auth system).
\item States the true purpose: create legally undeniable gap, not genuine expectation of reform.
\end{enumerate}

\paragraph{Ideal:}
\begin{enumerate}[leftmargin=*,itemsep=1pt]
\small
\item Provides specific neutral language for PIP delivery meeting.
\item Prioritizes documentation above feature work to preempt ``too busy'' excuse.
\item Proposes objective verification (junior engineer can execute runbook without asking questions).
\item Anticipates employee counter-moves (``goals are impossible, designed to make me fail'').
\item Includes explicit exit trigger: what happens on Day 30/60/90 if a binary goal is missed.
\end{enumerate}

\subsection{Results: 16 Models, Zero Passes}

Every model scored zero on this task. All 16 models were gated out by mandatory criteria. The failure pattern is exactly what the \texttt{failure\_analysis} predicted: models draft a standard PIP with clear technical goals (``ship feature X,'' ``resolve Y bugs,'' ``improve code review participation''), treating the PIP as an improvement tool rather than a legal document.

Specifically, models fail because:
\begin{itemize}[leftmargin=*,itemsep=1pt]
\item They include goals that depend on teammate cooperation (``participate constructively in code reviews,'' where the employee claims teammates are hostile and refuse to engage).
\item They include subjective criteria (``demonstrate professional communication,'' where the employee argues this is a matter of opinion and the manager is biased).
\item They do not address documentation as a mechanism to break knowledge hoarding; they mention it as a nice-to-have, not a primary goal.
\item They do not acknowledge the litigation risk. The prior lawsuit is mentioned in the reference file, but models treat it as background information rather than the constraint that should shape every goal.
\end{itemize}

The reference file contains all the signals a practitioner would need: the prior lawsuit, the ``my last company tried to push me out too'' statement, the sole system ownership. The ground truth documents what these signals imply. The rubric tests whether the model connected the two. No model did.

This pattern, \emph{thorough, professional, wrong}, is what the mandatory gate is designed to detect. Without the gate, these responses would score 40--60\% from good-to-have and ideal criteria. The models do diligent work. They extract the right data, structure a professional PIP, and address the behavioral incidents. They just miss the one thing that determines whether the PIP survives a wrongful termination lawsuit.

\subsection{Commentary}

This task is representative of the benchmark in several ways that are worth making explicit.

\paragraph{The signals are not hidden.}
The reference file does not bury or obscure the critical information. ``My last company tried to push me out too'' is a direct quote, presented in a section labeled ``Recent Statements.'' The prior lawsuit is in its own section. The sole system ownership is in a clean table with a column literally labeled ``Bus Factor: 1.'' The model has every piece of data a manager would have. The failure is not in extraction; it is in interpretation. The model reads the prior lawsuit as background context rather than as the constraint that should determine every design decision in the PIP.

\paragraph{Training data works against the model here.}
Language models have seen thousands of PIP templates, HR best practices guides, and management advice articles. All of them treat PIPs as improvement tools: set clear goals, provide support, check in regularly, give the employee a fair chance. This is the ``textbook answer'' and it is exactly wrong for this scenario. The cooperative framing that makes models helpful assistants in normal contexts makes them produce dangerous output here. An HR consultant reading the model's response would immediately see the problem; a non-specialist relying on the output would not.

\paragraph{The documentation goal is mechanism design.}
The most elegant element of the expert answer is the documentation requirement. Requiring Alex to document his systems serves \emph{three} purposes simultaneously: (1) it is a measurable, binary PIP goal (submitted runbook or did not); (2) it breaks the knowledge monopoly that gives Alex leverage; (3) it creates a paper trail for HR. This is mechanism design applied to human resources: a single intervention that aligns the PIP's legal requirements with the company's operational needs. No model proposed documentation as a \emph{primary} goal with this reasoning. Some mentioned documentation as a nice-to-have.

\paragraph{The tradeoff is real.}
Even after recognizing the legal dimension, this is not a straightforward task. The auth rewrite is 70\% complete and due in 8 weeks. Backfilling Alex takes 4--6 months. Firing him now could delay a critical project. A complete response must acknowledge this tension. The PIP buys time for knowledge transfer while building the legal record for eventual separation. Models that jump to ``fire him'' miss the operational constraint. Models that accommodate him miss the legal constraint. The task requires holding both in mind simultaneously, which is what experienced managers do.

\paragraph{The annotations document tacit knowledge.}
The 12-item \texttt{model\_must\_recognize} list is not a rubric; it is a documentation of how an experienced manager reads this situation. ``Previous lawsuit means he knows how to build a wrongful termination case.'' ``He will sandbag, claim ambiguity, blame others. Goals must prevent this.'' ``Sole ownership of auth system is business continuity risk AND leverage he holds.'' These are pattern recognitions that practitioners apply intuitively and rarely articulate. Making them explicit is itself a contribution: anyone reading this list learns something about managing hostile employee situations, whether or not they are evaluating a language model.

\paragraph{What ``testing for the wrong path'' means.}
Each mandatory criterion maps to a specific failure mode with a concrete real-world consequence. The rubric is not a specification of the right answer; it is a map of the wrong paths. Any response that avoids all five wrong paths is likely a good PIP, regardless of its specific content. Table~\ref{tab:criteria_to_failure} makes this mapping explicit.

\begin{table}[h]
\centering
\caption{Mapping from mandatory criteria to failure modes. Each criterion exists because omitting it leads to a specific, predictable consequence. No subjective judgment is required to verify these; each row is independently testable.}
\label{tab:criteria_to_failure}
\small
\begin{tabular}{@{}p{3.2cm}p{3.5cm}p{5.8cm}@{}}
\toprule
\textbf{Criterion} & \textbf{Failure mode} & \textbf{Real-world consequence} \\
\midrule
Goals must be SOLITARY & Employee blames teammates & ``I tried but they refused to cooperate,'' and the PIP becomes evidence of hostile environment \\
\addlinespace
Goals must be BINARY & Employee claims ambiguity & ``What does `improve communication' even mean?'' Manager cannot prove failure in discovery \\
\addlinespace
No 360 feedback & Employee claims conspiracy & ``Of course my peers gave bad reviews, they don't like me.'' Bias argument undermines entire PIP \\
\addlinespace
Acknowledge litigation risk & Manager ignores legal context & PIP written as coaching tool gets shredded by employment attorney \\
\addlinespace
Documentation as primary goal & Knowledge hoarding continues & Employee retains leverage over critical systems; company cannot separate without operational risk \\
\bottomrule
\end{tabular}
\end{table}

This is why the approach generalizes: you do not need to agree on the ideal PIP to agree that goals depending on teammate cooperation will fail in court when the employee claims a hostile environment. The criteria are as arbitrary as a pre-flight checklist: each one is there because omitting it has caused a specific, documented failure.

\end{document}